\PassOptionsToPackage{table}{xcolor}
\AddToHook{package/cleveref/before}{%
  \RequirePackage{amsmath,amssymb,amsthm}%
}
\documentclass{bytedance}

\usepackage[utf8]{inputenc}
\usepackage{wrapfig,array}
\usepackage{algorithm,algpseudocode}
\usepackage{url,xspace,bbding}
\newcommand{\method}{DeOverlay}
\newcommand{\R}{\mathbb{R}}
\newcommand{\E}{\mathbb{E}}
\newcommand{\vc}{\mathbf{c}}
\newcommand{\mask}{\widetilde M}

\title{DeOverlay: A Benchmark and Source-Anchored Flow for Video Overlay Removal}
\author{%
  \centerline{%
    \href{https://openreview.net/profile?id=~Deshui_Miao1}{Deshui Miao}$^{1}$\quad
    \href{https://openreview.net/profile?id=~Yuhang_Ma1}{Yuhang Ma}$^{1}$\quad
    \href{https://openreview.net/profile?id=~Shaoyong_Jia1}{Shaoyong Jia}$^{1}$\quad
    \href{https://openreview.net/profile?id=~Wenzhao_Gao1}{Wenzhao Gao}$^{1}$%
  }
  \centerline{%
    \href{https://openreview.net/profile?id=~Yue_Zhang40}{Yue Zhang}$^{1}$\quad
    \href{https://openreview.net/profile?id=~Shaohui_Jiao1}{Shaohui Jiao}$^{1}$\quad
    \href{https://openreview.net/profile?id=~Xin_Li24}{Xin Li}$^{2}$%
  }
}
\affiliation[1]{ByteDance}
\affiliation[2]{Southern University of Science and Technology}
\abstract{
Video overlay removal aims to recover scenes obscured by subtitles, titles, and logos while preserving surrounding content. Existing methods use temporal propagation or source-conditioned generation, but can leave residual overlays and distort scene structure when useful source information is obscured. We propose \textbf{\method{}}, a source-anchored flow model that reconstructs the clean video from the source latent with fixed source guidance. Specifically, we train a shared velocity network along paired source-to-clean paths using region-weighted flow matching, enabling successive corrections from intermediate states. To select local source cues, we develop visibility-normalized temporal aggregation that excludes masked positions and averages the remaining observations. Global source context and edit-region embeddings complement these cues through additive conditioning, retaining a single video-token sequence. To evaluate scene recovery across diverse overlay conditions, we introduce \textbf{DVTE-Bench}, comprising 1,631 paired videos across eight text-overlay categories, with OCR-screened clean targets, frame-aligned pixel masks, and a common evaluation protocol. Experiments on the complete release show that \method{} outperforms four baselines on all five reconstruction and perceptual metrics, reducing Mask LPIPS and Mask VFID by 58.3\% and 70.3\%, respectively, relative to the strongest evaluated baseline. Alongside these quantitative results, qualitative examples demonstrate the removal of graphic logos and stickers. To support further evaluation, we will make the complete benchmark, evaluation code, and test outputs available.
}

\checkdata[Project Page]{\url{https://yahooo-m.github.io/DVTE-Bench/}}

\hypersetup{
  pdfauthor={Deshui Miao, Yuhang Ma, Shaoyong Jia, Wenzhao Gao, Yue Zhang, Shaohui Jiao, Xin Li},
  pdftitle={DeOverlay: A Benchmark and Source-Anchored Flow for Video Overlay Removal}
}

\begin{document}
\maketitle
\suppressfloats[t]
\section{Introduction}

\begin{figure}[t]
\centering
\includegraphics[width=0.48\linewidth]{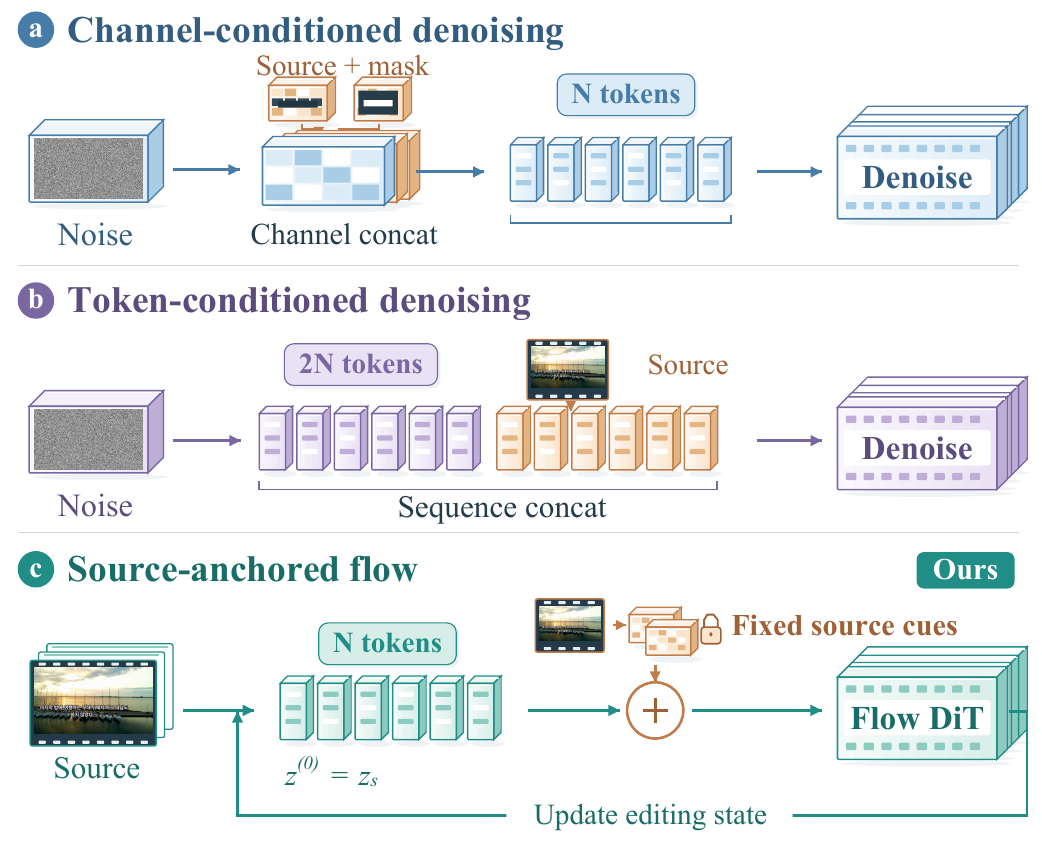}
\caption{\textbf{Starting state and source conditioning.} Channel- and sequence-conditioned editors start from noise, whereas ours starts from $z_s$ and receives fixed source features at each update. On a shared patch grid, the three designs use $N$, $2N$, and $N$ video tokens per call, with auxiliary tokens and codecs omitted from the diagram.}
\label{fig:mechanism_comparison}
\end{figure}

Video overlay removal aims to recover scene content obscured by subtitles, titles, watermarks, and logos while leaving the surrounding video unchanged. Although neighboring frames may reveal some occluded content, persistent coverage leaves other regions unobserved throughout the clip. Effective removal therefore requires both exploiting available observations and inferring missing content, while maintaining temporal coherence and preserving the scene structure already present in the input.

Propagation-based inpainting transfers image content and features across frames, but relies on learned priors when covered regions lack direct observations~\citep{zhou2023propainter}. Generative methods infer covered content through subtitle-specific reconstruction~\citep{he2026clear} or noise-initialized editing, in which source information enters through channel concatenation~\citep{zi2025minimax} or sequence concatenation~\citep{hui2026sedit} while the editable state begins from noise. Our paired comparisons show residual lettering and boundary distortions in some evaluated outputs (Figure~\ref{fig:benchmark_comparison}).

We propose \textbf{\method{}}, which starts reconstruction from the source latent and keeps the original source available to each update. A shared velocity network learns source-to-clean corrections from paired latent paths under a region-weighted flow objective, then applies successive corrections at inference as the state and time evolve under fixed weights and source guidance.

Visibility-normalized temporal aggregation averages unmasked source features across nearby frames, while a global source summary supplies scene context and an edit-region embedding marks the covered region. By adding these terms to the editing tokens at every update, we keep the source available without a second video-token sequence. Where no frame reveals the covered content, the video prior infers it.

Evaluating overlay removal requires checking whether the recovered content agrees with the underlying scene. Existing studies assess subtitle erasure~\citep{he2026clear} or visible watermark removal~\citep{chen2025deco} in separate task-specific settings. To support common regional evaluation across overlay conditions, we introduce \textbf{DVTE-Bench}, the Diverse Video Text Erasure Benchmark, with 1,631 paired videos covering eight text-overlay categories that vary in layout, persistence, and appearance. OCR-screened clean targets and aligned pixel masks provide the references for this evaluation. Experiments on the complete release show improvements over four baselines on all five reconstruction and perceptual metrics, while component and step-count studies examine the contribution of source guidance and repeated updates.

Our contributions are summarized as:
\begin{itemize}
\item We introduce \textbf{DVTE-Bench}, a paired benchmark covering eight text-overlay types with clean targets, aligned pixel masks, and common evaluation of regional reconstruction, perceptual quality, and temporal behavior.
\item We propose \textbf{\method{}}, which learns a source-to-clean flow with fixed local and global guidance. Source initialization retains observed structure, while visibility-normalized aggregation selects local cues for reconstructing covered regions.
\item We evaluate \method{} on all 1,631 released clips and obtain the best results among five methods on all five reconstruction and perceptual metrics. Ablations examine source guidance and flow step count, with separate measurements of inference cost.
\end{itemize}

\section{Related Work}
\paragraph{Video inpainting and overlay removal.}
ProPainter combines image and feature propagation with mask-guided sparse attention~\citep{zhou2023propainter}. For subtitle removal, CLEAR learns subtitle representations followed by generative reconstruction and supports mask-free inference~\citep{he2026clear}, while DECO combines frame-wise priors with temporal modeling for visible watermark removal~\citep{chen2025deco}. Our method learns source-to-clean corrections with persistent source guidance.

\paragraph{Source-conditioned restoration.}
SEDiT concatenates fixed source-reference tokens with noise-initialized tokens for one-step subtitle erasure~\citep{hui2026sedit}. For space-time super-resolution, DiffST combines input-derived initialization, motion-based frame aggregation, and video-prompt guidance~\citep{chen2026diffst}. We adapt its local/global decomposition through visibility-normalized aggregation at fixed coordinates and a pooled latent summary, which guide a source-initialized state through successive updates.

\paragraph{Flow-based editing.}
Flow matching and rectified flow learn vector fields along prescribed paths~\citep{lipman2023flow,liu2023rectified}. In restoration, IR-Flow learns degraded-to-clean image transport~\citep{fan2026irflow}, while PCFlow compares observation-initialized unconditional flow with noise-initialized conditional flow~\citep{jo2026pcflow}. For editing, FlowEdit performs inversion-free flow editing~\citep{kulikov2025flowedit}, and RFDM predicts residuals relative to preceding outputs for causal video editing~\citep{salehi2026rfdm}. Our paired video flow combines source initialization, fixed source guidance, and regional weighting within a pretrained video transformer~\citep{ma2026lingbot,robbyant2026code}.

\paragraph{Evaluation resources.}
CLEAR evaluates a 400-clip subtitle benchmark~\citep{he2026clear}, while SEDiT introduces VSR-Bench-400 with clean videos and subtitle masks~\citep{hui2026sedit}. In the watermark setting, DECO evaluates stationary and moving watermarks synthesized on YouTube-VOS and DAVIS~\citep{chen2025deco}.

\section{DVTE-Bench: Diverse Video Text Erasure}
\label{sec:benchmark}
We construct DVTE-Bench to evaluate overlay removal and scene reconstruction by pairing each overlaid video with a clean target and frame-aligned masks. These pairs support common evaluation across overlay layouts, temporal behaviors, and appearances.

Table~\ref{tab:benchmark_comparison} contrasts DVTE-Bench with existing subtitle benchmarks in paired supervision, overlay diversity, and evaluation coverage.
\begin{table}[t]
\centering
\caption{\textbf{Comparison of video text erasure benchmarks.} CLEAR~\citep{he2026clear} and VSR-Bench-400~\citep{hui2026sedit} are compared with DVTE-Bench. The DVTE-Bench release checkmark denotes its planned public release.}
\label{tab:benchmark_comparison}

\resizebox{\textwidth}{!}{%
\begin{tabular}{@{}lccccccccccc@{}}
\toprule

\multirow{2}{*}{Benchmark} &
\multirow{2}{*}{Clips} &
\multirow{2}{*}{Release} &
\multicolumn{2}{c}{Ground Truth} &
\multicolumn{5}{c}{Overlay Diversity} &
\multicolumn{2}{c}{Evaluation} \\

\cmidrule(lr){4-5}
\cmidrule(lr){6-10}
\cmidrule(lr){11-12}

& & &
Clean & Mask &
Types & Multilingual & Spatial & Temporal & Appearance &
Regional & Temporal \\

\midrule

CLEAR test set
& 400 & \XSolidBrush
& \Checkmark & \XSolidBrush
& 1 & \XSolidBrush & \Checkmark & \XSolidBrush & \Checkmark
& \XSolidBrush & \Checkmark \\

VSR-Bench-400
& 400 & \XSolidBrush
& \Checkmark & \Checkmark
& 1 & \Checkmark & \Checkmark & \XSolidBrush & \Checkmark
& \XSolidBrush & \XSolidBrush \\

\rowcolor{teal!8}
\textbf{DVTE-Bench (ours)}
& \textbf{1,631}
& \Checkmark
& \Checkmark & \Checkmark
& \textbf{8}
& \Checkmark & \Checkmark & \Checkmark & \Checkmark
& \Checkmark & \Checkmark \\

\bottomrule
\end{tabular}%
}
\end{table}

\subsection{Challenge Design}
\label{sec:benchmark_challenges}
The benchmark contains eight overlay types organized around three aspects of reconstruction (Table~\ref{tab:benchmark_coverage}). \emph{Spatial layout} determines which scene regions must be recovered, including vertical text, non-bottom placement, and multiple simultaneous regions. \emph{Temporal behavior} affects whether nearby frames expose the covered content, as illustrated by moving text and persistent watermarks. \emph{Appearance and coverage} vary through standard subtitles and artistic titles with display typography, outlines, or glow, while speech-aligned captions add audio-derived content and timing. Because these factors interact within each category, category comparisons measure coverage across conditions rather than isolated effects.

\begin{figure}[t]
\centering
\includegraphics[width=\linewidth]{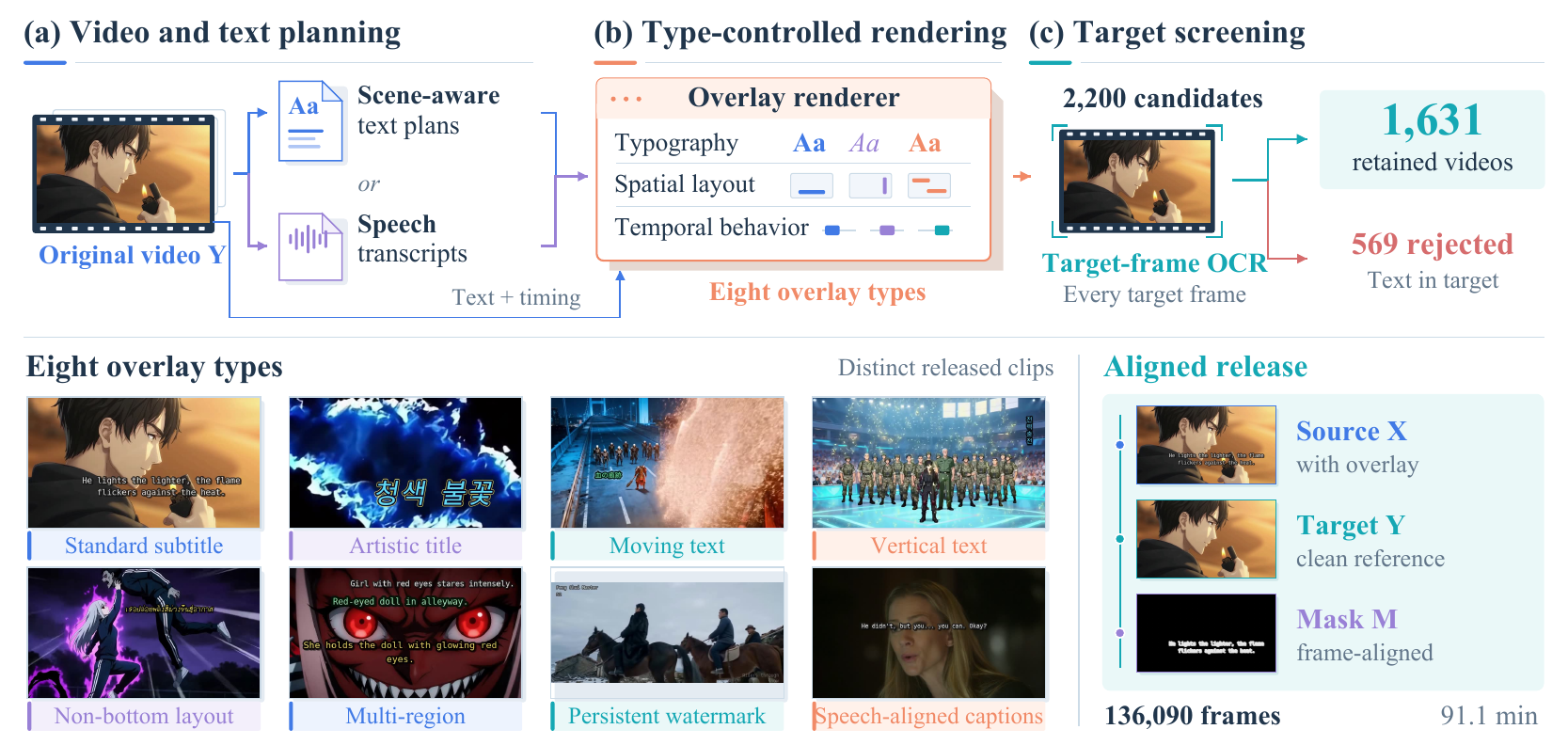}
\caption{\textbf{DVTE-Bench construction.} Text plans and type-controlled rendering produce aligned source videos, clean targets, and overlay masks. Target-frame OCR screening rejects 569 of 2,200 candidates and retains 1,631 clips. The eight thumbnails show distinct released clips, while the lower-right source, target, and mask form one aligned example.}
\label{fig:benchmark_construction}
\end{figure}

\subsection{Paired Data and Target Quality}
The release contains 1,631 clips, 136,090 frames, and 91.1 minutes of video, with 1,085 portrait and 546 landscape clips. As outlined in Figure~\ref{fig:benchmark_construction}, scene-aware text plans or speech transcripts supply overlay content for type-controlled rendering. This rendering retains the original video as the reconstruction target and produces aligned overlay masks, including masks that follow moving text. Language metadata has 11 labels, dominated by Chinese, English, Korean, and Japanese, with the released distributions shown in Figure~\ref{fig:benchmark_distribution}.

Because text in a reference can be mistaken for an unerased overlay, we screen every target frame with OCR and reject candidates with confirmed text, removing 569 of 2,200 candidates and retaining 1,631 (74.1\%). Preservation of natural scene text requires a separate evaluation setting.

\subsection{Regional Evaluation Protocol}
\label{sec:benchmark_protocol}
All five methods are evaluated on the same 1,631 released clips, with scoring based on pixel-accurate GT masks from the released NPZ archives independently of the bounding-box masks supplied to mask-conditioned models.

Mask PSNR measures pixel fidelity, and Mask SSIM~\citep{wang2004ssim} averages the SSIM map over GT mask pixels. Mask LPIPS~\citep{zhang2018lpips}, DISTS~\citep{ding2020dists}, and VFID measure perceptual agreement with clean targets. For these perceptual metrics, exterior prediction pixels are replaced by the target, and a shared crop covers the union of mask locations over time to focus evaluation on the overlay region. Frame measurements are averaged within videos and then equally across videos, whereas VFID is a global distance between video-feature distributions. Infinite-PSNR videos are omitted from the finite PSNR mean and counted as $N_{\infty}$.

Temporal evaluation reports mask-restricted TWE for predictions and clean targets, their aggregate gap, and auxiliary TC and flow statistics using the released Farneback protocol. These diagnostics complement target-referenced fidelity because temporally smooth outputs can still reconstruct the wrong content. Appendix~\ref{app:metrics} describes the regional computations and aggregation.

\section{DeOverlay: Source-Anchored Flow}
\label{sec:method}
\begin{figure}[t]
\centering
\includegraphics[width=\linewidth]{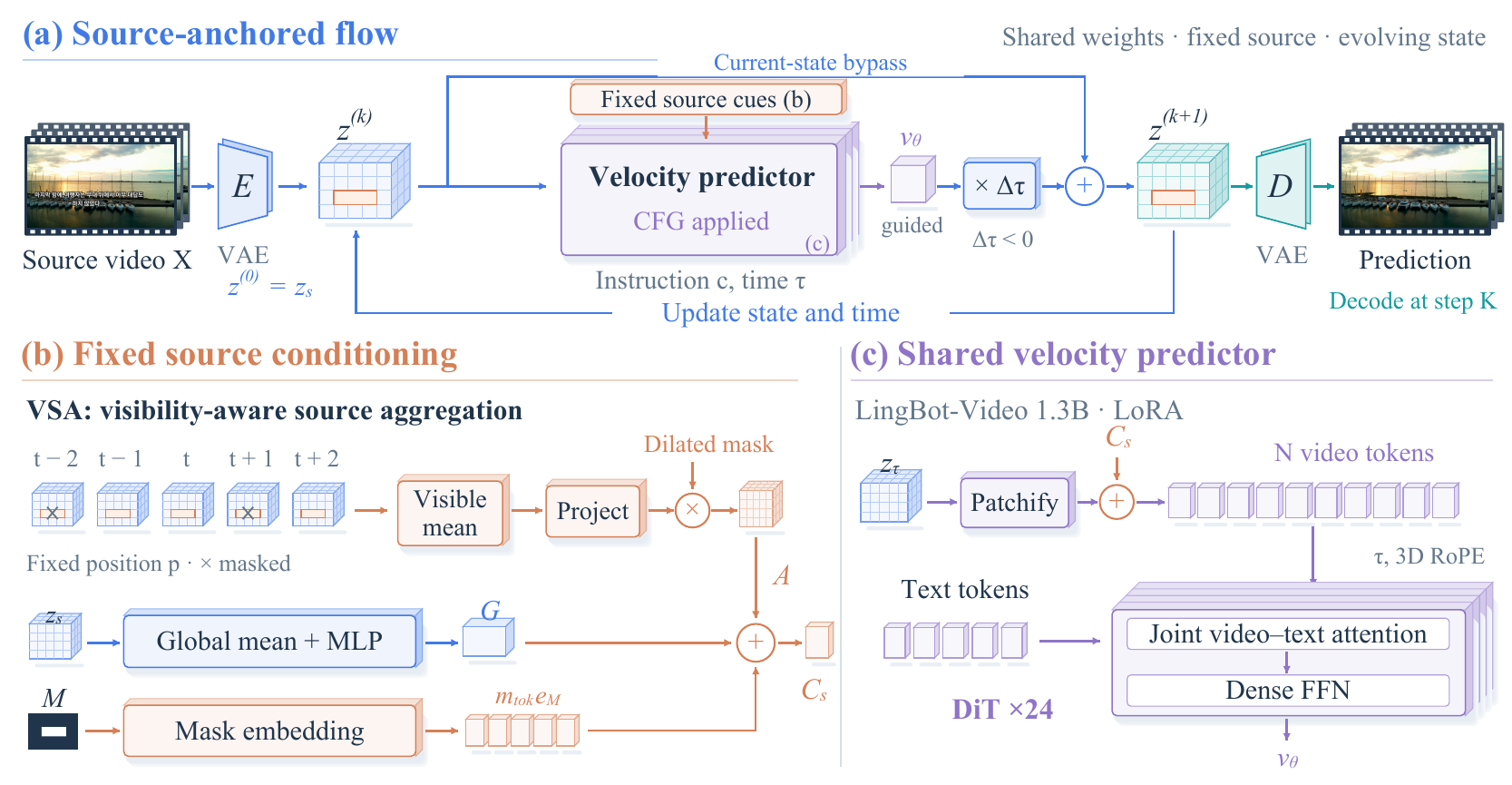}
\caption{\textbf{\method{} overview.} (a) Shared-weight Euler updates start from the source latent, with the upper bypass carrying the current state and the lower loop updating state and time with fixed source cues. (b) Visibility-aware source aggregation (VSA), global source conditioning (GSC), and mask embeddings form the source condition. (c) Source cues are added to the current state tokens before joint video--text processing. Tensor grids and intermediate states are schematic, whereas RGB frames are matched source/output pairs.}
\label{fig:overall}
\end{figure}

Given an overlaid video, a removal instruction, and an edit mask, we encode the source as the initial editing state and use a shared velocity network to predict corrections with fixed local and global source guidance. A video decoder then maps the updated latent to the reconstructed video (Figure~\ref{fig:overall}).

\subsection{Source-Initialized Continuous Flow}
To learn corrections from an overlaid video to its clean counterpart, we connect their latent representations with a paired flow path~\citep{lipman2023flow,liu2023rectified}. Let $X,Y\in\R^{T\times H\times W\times3}$ denote the aligned source and target, with $M$ marking affected pixels, including partially transparent regions. A video VAE encodes $z_s=\mathcal E(X)$ and $z_c=\mathcal E(Y)$ in $\R^{C\times T_\ell\times H_\ell\times W_\ell}$, with $\mathcal D$ denoting decoding and $\vc$ the removal instruction. The training path is
\begin{equation}
 a=z_s+\sigma\epsilon,\qquad z_\tau=(1-\tau)z_c+\tau a,
 \qquad \epsilon\sim\mathcal N(0,I),\quad \tau\in[0,1],
 \label{eq:path}
\end{equation}
with $\sigma=0.05$, the source endpoint at $\tau=1$, and the clean endpoint at $\tau=0$, giving the target velocity
\begin{equation}
 v^*=\partial z_\tau/\partial\tau=a-z_c.
 \label{eq:velocity}
\end{equation}
We sample intermediate $\tau$ so that the predictor $v_\theta(z_\tau,\tau,\vc;z_s,M_{\rm model})$ learns corrections at different positions along the path from the current state and time together with the original source and conditioning mask. The perturbation augments training, while inference starts directly at $z_s$.

\subsection{Persistent Source Guidance}
\label{sec:guidance}
As editing changes the reconstruction, fixed source guidance keeps the original observations available. We combine local source features, a global summary, and the edit mask with the patchified state by addition:
\begin{equation}
 h_\tau=P(z_\tau)+A(z_s,M_{\rm model})+G(z_s)+m_{\rm tok}e_M.
 \label{eq:fusion}
\end{equation}
Here $P$ embeds the state, $m_{\rm tok}$ is the token-grid mask, and $e_M$ is a learned vector. All terms have the transformer width, with $G$ broadcast over tokens, and their addition preserves the number of video tokens.

\paragraph{Visibility-aware source aggregation.}
Direct temporal averaging can mix overlay features with useful scene information. Inspired by the local/global decomposition in DiffST~\citep{chen2026diffst}, our visibility-aware source aggregation (VSA) excludes masked positions and normalizes by the number of visible observations. Let $s_{t,p}$ be the source latent average-pooled to the token grid and $m_{t,p}$ its max-pooled mask. Within $\mathcal N(t)=\{t-2,\ldots,t+2\}$, we compute
\begin{equation}
 \bar s_{t,p}=
 \frac{\frac{1}{5}\sum_{j\in\mathcal N(t)}(1-m_{j,p})s_{j,p}}
 {\max\!\left(\frac{1}{5}\sum_{j\in\mathcal N(t)}(1-m_{j,p}),\delta\right)},
 \qquad \delta=10^{-4}.
 \label{eq:local}
\end{equation}
The denominator prevents masked positions from attenuating the visible-feature mean, and we project $\bar s$ to transformer width with a two-layer pointwise network before gating it to the edit region with a dilated mask. Video attention supplies broader context, but fixed-coordinate aggregation can mix structures under motion and provides no local observations in fully masked windows.

\paragraph{Global source conditioning.}
To provide scene-level context when persistent coverage leaves local observations unavailable, our global source conditioning (GSC) projects the source's spatiotemporal mean over both overlaid and uncovered positions through a two-layer MLP. The resulting summary $G$ complements local features, while $m_{\rm tok}e_M$ marks the edit region. Output projections and $e_M$ start at zero, with further implementation details in Appendix~\ref{app:implementation}.

\subsection{Learning the Velocity Field}
\label{sec:training}
To emphasize overlay correction, we assign larger loss weights to covered regions. Let $\mask$ be the annotated mask resized to the latent grid and dilated by one cell. With $q_i$ excluding overlap-conditioning elements, we minimize
\begin{equation}
 \mathcal L_{\rm FM}=\E\!\left[
 \frac{\sum_i q_i(1+\lambda\mask_i)(v_{\theta,i}-v_i^*)^2}
 {\max(\sum_i q_i,1)}\right],\qquad \lambda=5.
 \label{eq:loss}
\end{equation}
The index spans channel, time, space, and batch, with covered elements receiving weight six and exterior elements weight one, while the denominator counts supervised elements without weighting.

We adapt the dense 1.3B LingBot-Video backbone~\citep{ma2026lingbot,robbyant2026code} with rank-256 LoRA~\citep{hu2022lora}, directly training the patch embedding and source-guidance modules while freezing the VAE and text encoder. Conditioned video tokens and projected instruction tokens undergo joint attention. With probability $0.2$, missing-mask training sets $M_{\rm model}=0$, zeroing local guidance and the edit embedding while retaining global guidance. The annotated mask continues to determine loss weights, and Table~\ref{tab:config} records the full configuration.

\subsection{Shared-Weight Flow Updates}
\label{sec:inference}
Starting at $z^{(0)}=z_s$, we use a descending schedule $1=\tau_0>\cdots>\tau_K=0$:
\begin{equation}
 z^{(k+1)}=z^{(k)}+\Delta\tau_k\,
 \widehat v_\theta(z^{(k)},\tau_k,\vc;z_s,M_{\rm model}),
 \qquad \Delta\tau_k=\tau_{k+1}-\tau_k<0.
 \label{eq:euler}
\end{equation}
The state and time change while weights, source, and mask remain fixed, so one checkpoint supports multiple step counts. We use a linear--quadratic schedule with a resolution-dependent shift and apply classifier-free guidance~\citep{ho2022cfg} at scale $g=2$ with two network evaluations per step, so six steps require 12 evaluations. The core output is $\widehat Y=\mathcal D(z^{(K)})$, with chunked inference and output processing specified in Appendix~\ref{app:implementation}.

Although each training pair defines a straight latent path, the learned velocity can vary with the state and time during inference. We therefore evaluate how the number of flow updates affects reconstruction quality in Section~\ref{sec:step_study}. Appendix~\ref{app:derivation} analyzes the integration and endpoint errors.

% Keep the experiment floats close to their discussion without stretching gaps.
\begingroup
\raggedbottom
\setlength{\floatsep}{9pt plus 1pt minus 1pt}
\setlength{\textfloatsep}{11pt plus 1pt minus 1pt}
\setlength{\intextsep}{10pt plus 1pt minus 1pt}
\setlength{\abovecaptionskip}{6pt}
\section{Experiments}
\label{sec:experiments}

\subsection{Experimental Setup}
\paragraph{Training, inference, and baselines.}
Training uses 350,000 aligned overlaid/clean clip pairs with masks: 250,000 base samples, 50,000 stylized-text samples, and 50,000 mirrored-text samples. These clips are disjoint from DVTE-Bench, with corpus construction and optimization settings detailed in Appendix~\ref{app:implementation}. Full-release evaluation uses the 40K checkpoint, GT bounding-box input masks, $K=6$, guidance scale $g=2$ (12 network evaluations), and 12-pixel seamless paste-back. We compare against ProPainter~\citep{zhou2023propainter}, CLEAR~\citep{he2026clear}, MiniMax-Remover~\citep{zi2025minimax}, and DiffuEraser~\citep{li2025diffueraser}, spanning propagation, subtitle-specific, and generative removal, with non-subtitle cases testing CLEAR's transfer beyond its original task. All methods share GT scoring masks, while input-mask access and postprocessing follow their released pipelines (Appendix~\ref{app:categories}). SEDiT~\citep{hui2026sedit} is compared conceptually because an executable model was unavailable for our matched evaluation. Ablations use 200 randomly sampled videos from DVTE-Bench.

\subsection{Reconstruction across Diverse Overlays}
\label{sec:main_results}
On all 1,631 released videos, \method{} achieves the best results on all five reconstruction and perceptual metrics (Table~\ref{tab:main}). Compared with MiniMax-Remover, the strongest baseline on these metrics, it improves Mask PSNR from 26.17 to 30.80 dB and SSIM from 0.8044 to 0.8822, while reducing LPIPS, DISTS, and VFID by 58.3\%, 40.8\%, and 70.3\%, respectively. These improvements support more faithful covered-region reconstruction.

\begin{table}[t]
\caption{\textbf{DVTE-Bench results on all 1,631 videos}, using the released pixel-mask protocol. (a) PSNR$^{\dagger}$ excludes infinite values, and VFID is a global video-feature distance. (b) TWE Gap is $|\overline{\mathrm{TWE}}_{\rm pred}-\overline{\mathrm{TWE}}_{\rm GT}|$, with clean-target TWE 9.3228 and diagnostics defined in Appendix~\ref{app:temporal_diagnostics}. Bold marks best values in (a), while bold/underline in (b) mark the closest/second-closest TWE Pred to GT and the lowest/second-lowest values otherwise.}
\label{tab:main}
\centering\small
\setlength{\tabcolsep}{4pt}
\begin{tabular*}{\linewidth}{@{\extracolsep{\fill}}lrrrrr@{}}
\toprule
\multicolumn{6}{l}{\textit{(a) Regional reconstruction and perceptual quality}}\\
Method & PSNR$^{\dagger}\uparrow$ & SSIM$\uparrow$ & LPIPS$\downarrow$ & DISTS$\downarrow$ & VFID$\downarrow$\\
\midrule
CLEAR & 16.02 & 0.5172 & 0.3209 & 0.2158 & 1.4736\\
ProPainter & 25.90 & 0.7832 & 0.1166 & 0.0957 & 0.3879\\
MiniMax-Remover & 26.17 & 0.8044 & 0.0977 & 0.0856 & 0.2806\\
DiffuEraser & 24.84 & 0.7723 & 0.1130 & 0.0952 & 0.3285\\
\method{}, $K=6$ & \textbf{30.80} & \textbf{0.8822} & \textbf{0.0408} & \textbf{0.0507} & \textbf{0.0835}\\
\bottomrule
\end{tabular*}
\par\smallskip
\begin{tabular*}{\linewidth}{@{\extracolsep{\fill}}lrrrrr@{}}
\toprule
\multicolumn{6}{l}{\textit{(b) Regional temporal diagnostics}}\\
Method & TWE Pred & TWE Gap & TC & Flow Mean & Flow Var\\
\midrule
CLEAR & 10.2373 & 0.9144 & 11.0261 & 0.9014 & 2.8169\\
ProPainter & 8.1428 & 1.1801 & \textbf{4.0349} & \underline{0.6496} & \underline{1.5601}\\
MiniMax-Remover & \underline{8.7095} & \underline{0.6133} & 4.5586 & 0.7991 & 2.3072\\
DiffuEraser & 10.1853 & 0.8624 & 5.3240 & 0.6690 & 1.9385\\
\method{} & \textbf{9.3110} & \textbf{0.0118} & \underline{4.2498} & \textbf{0.3333} & \textbf{0.5921}\\
\bottomrule
\end{tabular*}
\end{table}

\begin{figure}[!t]
\centering
\includegraphics[page=1,width=\linewidth]{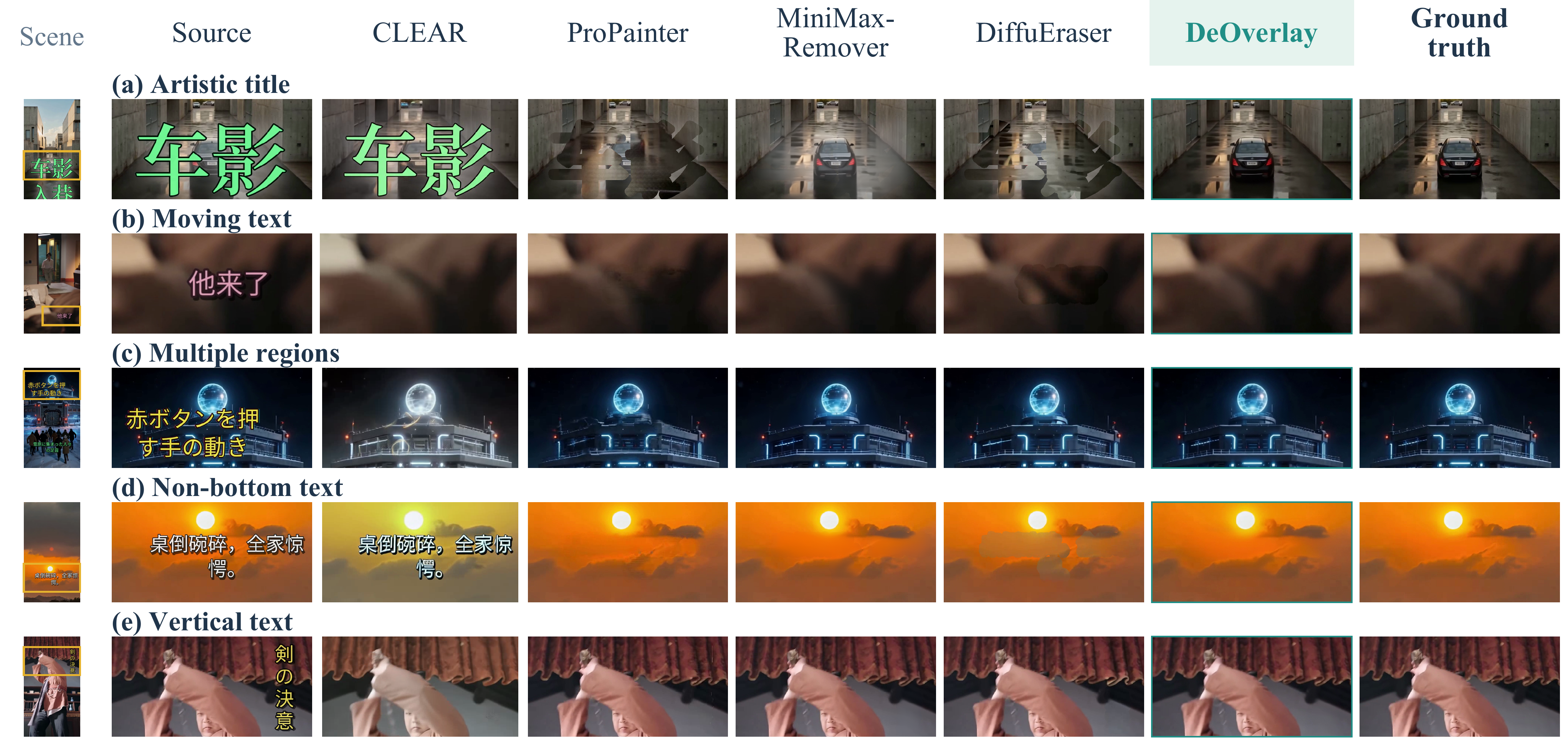}
\caption{\textbf{Paired benchmark comparisons.} Five frame-40 examples with shared yellow crop boxes compare four baselines, \method{}, and clean targets, with inference settings and expanded seven-category comparisons in Appendix~\ref{app:qualitative_exports}.}
\label{fig:benchmark_comparison}
\end{figure}

\paragraph{Coverage across overlay conditions.}
\method{} achieves the best PSNR, SSIM, LPIPS, and DISTS across all eight overlay types (Appendix~\ref{app:categories}). Artistic titles produce its lowest PSNR/SSIM (26.24 dB/0.8409), whereas vertical text produces its highest LPIPS/DISTS (0.0975/0.1015). The most difficult category therefore depends on whether pixel accuracy or perceptual similarity is measured.

\paragraph{Temporal behavior.}
Table~\ref{tab:main}(b) reports the released temporal diagnostics alongside reconstruction fidelity, with \method{} obtaining TWE Pred of 9.3110 and TC of 4.2498, compared with ProPainter's 8.1428 and 4.0349. All methods use the same mask-restricted computation and clean-target reference of 9.3228. Under this computation, ProPainter and MiniMax-Remover score below the reference despite weaker reconstruction, illustrating why the diagnostic is interpreted together with clean-target fidelity. Appendix~\ref{app:temporal_diagnostics} specifies the temporal computations.

\method{} has an aggregate TWE Gap of 0.0118, versus 0.6133 for MiniMax-Remover and 1.1801 for ProPainter, where the gap compares dataset-level means under the aggregation rule in Appendix~\ref{app:temporal_diagnostics}. The five target-referenced metrics in Table~\ref{tab:main}(a) therefore provide the primary reconstruction comparison, while Flow Mean, Flow Var, and TC retain their released labels and scales as supplementary descriptions of the evaluated outputs.

\paragraph{Qualitative reconstruction.}
Across the five paired examples with clean targets in Figure~\ref{fig:benchmark_comparison}, residual text, blurred structures, and altered colors distinguish the outputs, while \method{} more closely preserves the target appearance. In the artistic-title example, the restored car and road remain aligned with the clean target. The multiple-region and non-bottom-text examples also make color shifts and damaged background boundaries visible around the building and clouds, complementing the aggregate scores by showing which structures are recovered within the covered regions. Figure~\ref{fig:qualitative} adds twelve application examples spanning subtitle languages and density, title layouts, watermarks, a sticker, and a television logo. Appendix~\ref{app:qualitative_exports} provides seven-category comparisons, application full frames, and additional successes and failures.

\begin{figure}[!t]
\centering
\includegraphics[width=\linewidth]{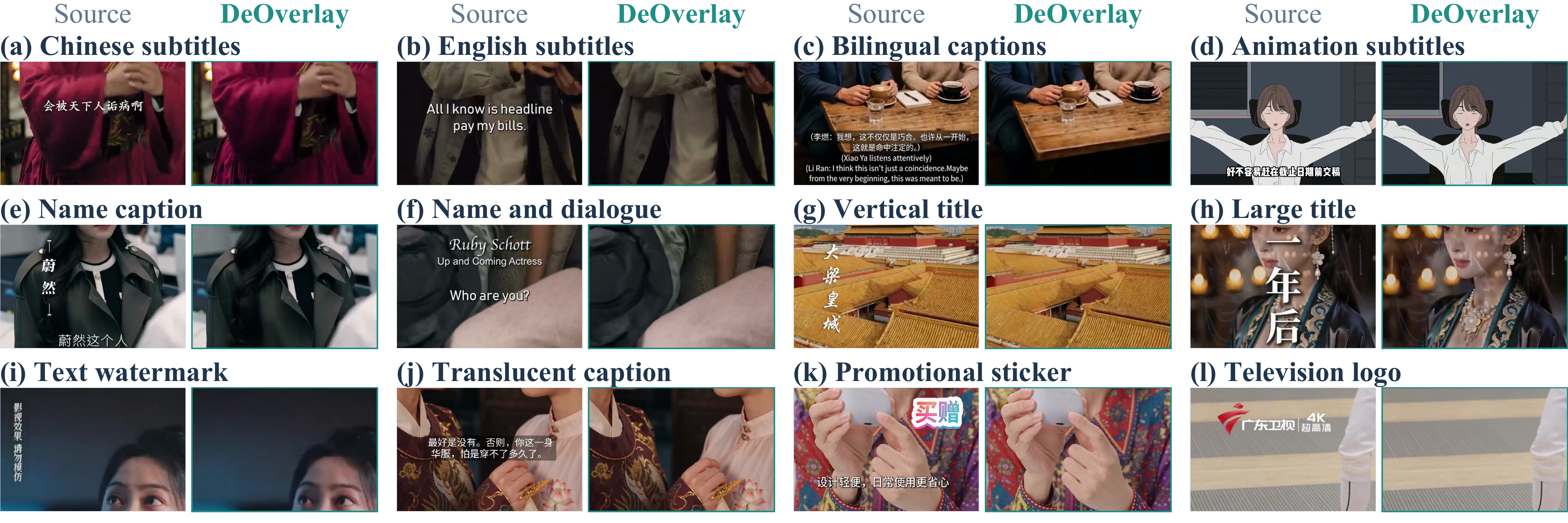}
\caption{\textbf{Removal across overlay appearances.} Twelve matched source/output crops cover subtitles, titles, watermarks, a sticker, and a television logo, with full frames and crop boxes in Appendix~\ref{app:qualitative_exports}.}
\label{fig:qualitative}
\end{figure}

\subsection{Ablation Studies}
\label{sec:step_study}
Ablations use 200 randomly sampled DVTE-Bench videos, whereas Table~\ref{tab:main} uses all 1,631 videos. Component variants use $K=4$, with a direct-residual reference at $K=1$, and the step sweep holds the checkpoint fixed. We measure ablation SSIM with MaskCrop SSIM.

\paragraph{Effect of source representation and guidance.}
Replacing source initialization and additive guidance with noise initialization and concatenated source tokens decreases Mask PSNR by 2.861 dB (Table~\ref{tab:core_ablation}). Because the starting state and conditioning change together, this comparison evaluates their combined design.

Within the source-initialized model, keeping only the mask embedding reduces PSNR by 0.231 dB, while removing VSA or GSC reduces it by 0.091 or 0.171 dB. These changes show that both local observations and global context contribute to the measured reconstruction quality. Replacing visibility-normalized aggregation with an ordinary temporal mean reduces PSNR by 0.101 dB and raises MAE to 4.360, supporting the use of visibility selection in local aggregation.

\begin{table}[t]
\caption{\textbf{Core design ablations on 200 randomly sampled videos.} Component variants use $K=4$, and the direct-residual reference uses $K=1$. PSNR and MAE use overlay pixels, with $\Delta$PSNR measured relative to the full four-step model. VSA supplies local source context and GSC supplies global context, while the noise-start variant concatenates noise and source video tokens. Bold marks the best value in each quality column.}
\label{tab:core_ablation}
\centering\footnotesize
\setlength{\tabcolsep}{3pt}
\begin{tabular*}{\linewidth}{@{\extracolsep{\fill}}lrrrrrr@{}}
\toprule
Variant & PSNR$\uparrow$ & MAE$\downarrow$ & Crop SSIM$\uparrow$ & Outside MAE$\downarrow$ & TWE gap$\downarrow$ & $\Delta$PSNR\\
\midrule
\method{}, $K=4$ & \textbf{32.461} & \textbf{4.299} & \textbf{0.92765} & \textbf{2.541} & \textbf{0.180} & 0.000\\
Direct residual, $K=1$ & 31.300 & 4.520 & 0.91120 & 2.548 & 0.195 & -1.161\\
Noise init. + source tokens & 29.600 & 6.500 & 0.89200 & 3.050 & 0.320 & -2.861\\
Mask only (no VSA/GSC) & 32.230 & 4.440 & 0.92590 & 2.552 & 0.197 & -0.231\\
Ordinary temporal mean & 32.360 & 4.360 & 0.92725 & 2.542 & 0.183 & -0.101\\
Without VSA & 32.370 & 4.350 & 0.92705 & 2.543 & 0.185 & -0.091\\
Without GSC & 32.290 & 4.410 & 0.92645 & 2.550 & 0.193 & -0.171\\
\bottomrule
\end{tabular*}
\par\smallskip
\begin{minipage}{\linewidth}\footnotesize
Crop SSIM denotes MaskCrop SSIM, and TWE gap denotes the TWE absolute gap.
\end{minipage}
\end{table}

\paragraph{Effect of flow step count.}
Increasing from two to four steps within the same checkpoint improves Mask PSNR from 32.410 to 32.461 dB, lowers MAE from 4.330 to 4.299, and improves Crop SSIM and TWE gap (Table~\ref{tab:steps}). Eight steps add only 0.007 dB over four at $1.50\times$ latency, favoring four steps for these component studies.

\begin{table}[!tb]
\caption{\textbf{Flow step count on 200 randomly sampled videos.} Only the inference update count changes under a fixed checkpoint, with latency normalized to $K=4$ and the best quality metric marked in bold.}
\label{tab:steps}
\centering\small
\setlength{\tabcolsep}{4pt}
\begin{tabular*}{\linewidth}{@{\extracolsep{\fill}}crrrrr@{}}
\toprule
$K$ & Mask PSNR$\uparrow$ & Mask MAE$\downarrow$ & MaskCrop SSIM$\uparrow$ & TWE abs. gap$\downarrow$ & Rel. latency\\
\midrule
2 & 32.410 & 4.330 & 0.92710 & 0.184 & $0.82\times$\\
4 & 32.461 & 4.299 & 0.92765 & 0.180 & $1.00\times$\\
6 & 32.467 & 4.294 & 0.92772 & \textbf{0.178} & $1.25\times$\\
8 & \textbf{32.468} & \textbf{4.293} & \textbf{0.92774} & \textbf{0.178} & $1.50\times$\\
\bottomrule
\end{tabular*}
\end{table}

\paragraph{Training formulation and repeated updates.}
Four-step flow exceeds the separately trained direct-residual reference by 1.161 dB in Mask PSNR, with lower MAE and higher Crop SSIM (Table~\ref{tab:core_ablation}). Each flow update recomputes the velocity from the current state and time with fixed source cues, allowing different inference budgets without retraining.

\subsection{Efficiency}
\label{sec:efficiency}
\paragraph{End-to-end speed.}
The comparison in Table~\ref{tab:runtime} uses eight shared clips (650 frames), run serially on one A100, with timing that includes decoding, postprocessing, and encoding but excludes model loading. One-step \method{} reaches 3.0847 FPS, $1.17\times$ MiniMax-Remover's throughput, while four steps reach 2.0712 FPS. Both configurations have a sampled peak memory of 12.236 GiB, compared with the lowest value of 10.776 GiB for MiniMax-Remover. Appendix~\ref{app:reporting_details} specifies the measurement boundaries.

\begin{table}[!tb]
\caption{\textbf{Speed and memory on the same A100.} All rows process the same eight clips (650 frames), with E2E denoting end-to-end time excluding model initialization. Inference time follows each runner's CUDA-synchronized interval, and peak memory uses a common external GPU-memory sampler. Bold marks the best value in each column.}
\label{tab:runtime}
\centering\small
\setlength{\tabcolsep}{4pt}
\begin{tabular*}{\linewidth}{@{\extracolsep{\fill}}lrrrr@{}}
\toprule
Method & E2E s/frame$\downarrow$ & E2E FPS$\uparrow$ & Inference s/frame$\downarrow$ & Peak GiB$\downarrow$\\
\midrule
CLEAR & 0.6030 & 1.6584 & 0.5983 & 18.354\\
ProPainter & 0.4271 & 2.3413 & 0.4134 & 34.101\\
MiniMax-Remover & 0.3787 & 2.6405 & 0.3530 & \textbf{10.776}\\
DiffuEraser & 1.1677 & 0.8564 & 1.1524 & 23.950\\
\method{}, $K=1$ & \textbf{0.3242} & \textbf{3.0847} & \textbf{0.3001} & 12.236\\
\method{}, $K=4$ & 0.4828 & 2.0712 & 0.4587 & 12.236\\
\bottomrule
\end{tabular*}
\end{table}

\paragraph{Cost of a second token stream.}
With the same 32K weights and four transformer forwards, adding a source-token stream increases tokens from 13,346 to 26,594, median time by $2.485\times$, and incremental peak memory by $1.995\times$ (Table~\ref{tab:token_cost}). These measurements isolate transformer cost, whereas end-to-end timing also includes video-processing overhead. Additive guidance preserves the video sequence length, while increasing $K$ repeats the predictor.

\section{Limitations}
DVTE-Bench focuses on synthetic text overlays, leaving graphic logos, stickers, and preservation of natural scene text for broader paired evaluation. When neighboring frames provide no visible evidence for a fully covered background, reconstruction depends on the video prior. The application examples also include residual graphics and local texture artifacts (Appendix~\ref{app:qualitative_exports}).

\section{Conclusion}
We introduced DVTE-Bench and \method{} for reconstructing scenes beneath video text overlays, with DVTE-Bench unifying eight overlay categories in 1,631 paired videos. Clean targets, aligned pixel masks, and a common evaluation protocol support comparisons across overlay layouts, temporal behavior, and appearance.

\method{} combines source initialization with visibility-normalized local aggregation and global source guidance, keeping the source available through additive conditioning throughout shared-weight flow updates using one video-token sequence. It outperforms four baselines on all five reconstruction and perceptual metrics, and on the 200-video subset, four-step flow exceeds the separately trained direct-residual reference by 1.161 dB in Mask PSNR. Component ablations support both guidance branches, while category-level evaluation separates the challenges of pixel reconstruction and perceptual quality, with artistic titles and vertical text stressing different metrics. Together, the benchmark and method enable joint study of scene recovery and inference cost.

\endgroup
\FloatBarrier
\bibliographystyle{plainnat}
\bibliography{references}
\clearpage
\appendix
\section*{Appendix Contents}
\begingroup
\small
\setlength{\parskip}{2pt}
\newcommand{\appendixcontentsentry}[2]{%
  \noindent\hyperref[#1]{\ref*{#1}\enspace #2}\dotfill
  \hyperref[#1]{\pageref*{#1}}\par}
\appendixcontentsentry{app:implementation}{Implementation Details}
\appendixcontentsentry{app:derivation}{Euler Error and Endpoint Prediction}
\appendixcontentsentry{app:metrics}{Metric Definitions and Reporting Protocol}
\appendixcontentsentry{app:categories}{Overlay Categories and Evaluation Coverage}
\appendixcontentsentry{app:qualitative_exports}{Qualitative Exports and Case Analysis}
\appendixcontentsentry{app:mechanism_studies}{Ablation Details}
\appendixcontentsentry{app:reporting_details}{Timing and Optimization Protocol}
\appendixcontentsentry{app:category_comparison}{Complete Per-Category Comparison}
\appendixcontentsentry{app:temporal_diagnostics}{Temporal Diagnostic Computation}
\endgroup

\section{Implementation Details}
\label{app:implementation}
\paragraph{Training corpus.}
The training corpus contains 350,000 paired samples: 250,000 base samples, 50,000 stylized-text samples, and 50,000 mirrored-text samples, each providing an overlaid input, its aligned clean target, and an overlay mask. All training samples are separate from the 1,631 DVTE-Bench clips, including its 68 speech-aligned subtitle clips. We sample the base, stylized, and mirrored subsets with weights $4:2:1$, respectively, which specify sampling frequencies rather than corpus sizes.

Table~\ref{tab:config} gives the multistep configuration.

\begin{table}[h]
\caption{Configured model and optimization settings.}
\label{tab:config}
\centering\small
\begin{tabular}{ll}
\toprule
Setting & Value\\
\midrule
Backbone & LingBot-Video dense 1.3B configuration\\
Latent channels / patch size & 16 / $(1,2,2)$\\
Transformer blocks / width / heads & 24 / 2048 / 16\\
Feed-forward width / text embedding width & 6144 / 2560\\
LoRA rank / scale & 256 / 1.0\\
Local / global branch hidden width & 512 / 512\\
Local temporal window & 5 latent positions\\
Local mask dilation radius & 1 token-grid cell\\
Loss-mask dilation radius & 1 latent-grid cell\\
Model-mask dropout probability & 0.2\\
Source-endpoint perturbation $\sigma$ & 0.05 (training)\\
Overlay-region loss coefficient $\lambda$ & 5.0\\
Pixel-crop / high-frequency loss weights & 0 / 0\\
Optimizer / learning rate & AdamW / $5\times10^{-5}$\\
Betas / weight decay & $(0.9,0.95)$ / 0\\
Warm-up / maximum configured updates & 300 / 60,000\\
Precision / gradient clipping & bfloat16 / norm 1.0\\
Preview steps / batch inference default & 4 / 6\\
DVTE-Bench checkpoint / mask / steps & 40K / GT / 6\\
Guidance scale & 2.0\\
VAE posterior mode & Sample\\
\bottomrule
\end{tabular}
\end{table}

\paragraph{Token-grid conditioning.}
Before local context aggregation, the transformer averages the source latent over each spatial patch and max-pools the model mask to the same grid, with a temporal patch size of one. The local branch therefore operates on patch-grid source features, while the global branch averages the original latent over all temporal and spatial positions. Accordingly, mask dilation in the local branch and dilation in the loss use their respective resolutions.

\paragraph{Source-guidance projections.}
The local projection uses two $1\times1\times1$ convolutions with hidden width 512:
\begin{equation}
 A_{t,p}=\operatorname{dilate}(m)_{t,p}\,
 W_{A,2}\!\left(\operatorname{SiLU}(W_{A,1}\bar s_{t,p}+b_{A,1})\right)
 +\operatorname{dilate}(m)_{t,p}\,b_{A,2}.
 \label{eq:localproj}
\end{equation}
Dilation has radius one on the token grid, and padded temporal positions contribute zero to both sums in Eq.~\ref{eq:local}. The global source branch uses a two-layer MLP with hidden width 512:
\begin{equation}
 G(z_s)=W_{G,2}\operatorname{SiLU}\!\left(W_{G,1}
       \frac{1}{T_\ell H_\ell W_\ell}\sum_{t,p}z_{s,t,p}+b_{G,1}\right)+b_{G,2}.
 \label{eq:global}
\end{equation}
Both guidance output projections and the mask embedding are initialized to zero, and joint video--text self-attention uses three-dimensional rotary embeddings, time-dependent modulation, and dense feed-forward layers. LoRA of rank 256 adapts eligible linear layers, while the patch embedding, guidance modules, and mask embedding are trained directly. The VAE and text encoder remain fixed, with posterior sampling used in the VAE and zero weight assigned to pixel-crop and high-frequency losses.

\paragraph{Guidance and deployment.}
For classifier-free guidance~\citep{ho2022cfg}, positive and negative text conditions give
\begin{equation}
 \widehat v_\theta=v_\theta^-+g(v_\theta^+-v_\theta^-).
 \label{eq:cfg}
\end{equation}
With the source and conditioning mask retained in both evaluations, the default $g=2$ requires two network evaluations per Euler step. The decoder maps the final latent to the output video, and long videos use overlapping chunks in which previous outputs supply latent boundary conditions. Deployment derives text masks from OCR, while removing non-text overlays requires a mask covering the designated graphic region. Full-benchmark evaluation uses 12-pixel seamless pixel paste-back, with the method-specific output processing specified in Appendix~\ref{app:categories}.

\paragraph{Resolution and time sampling.}
The configured data buckets use 41 frames at a short-side resolution of 768 for videos with original long side at least 1920, 65 frames at 768 for long side at least 1280, and 121 frames at 608 for smaller inputs. The decoder samples a random starting frame with unit frame stride. The training sampler uses the continuous shifted-mode distribution with mode scale 1.1. The configured resolution shift ranges from 1 to 5 over the selected token-count range, and inference also applies the resolution shift to the linear--quadratic schedule.

\paragraph{Training and inference procedure.}
Algorithm~\ref{alg:flow} describes one training draw and core inference, omitting batch padding and chunk overlap for clarity. When overlap conditioning is active, the corresponding latent elements are supplied as conditions and excluded from the target mask $q$ in Eq.~\ref{eq:loss}.

\begin{algorithm}[h]
\caption{\method{} training and inference}
\label{alg:flow}
\begin{algorithmic}[1]
\Require Paired $X,Y,M$, instruction $\vc$, velocity network $v_\theta$
\Statex \textbf{Training}
\State Encode $z_s\gets\mathcal E(X)$ and $z_c\gets\mathcal E(Y)$
\State Sample $\epsilon\sim\mathcal N(0,I)$ and continuous time $\tau$
\State $a\gets z_s+0.05\epsilon$; $z_\tau\gets(1-\tau)z_c+\tau a$
\State Set $M_{\rm model}\gets0$ with probability $0.2$, otherwise $M$
\State Predict $v\gets v_\theta(z_\tau,\tau,\vc;z_s,M_{\rm model})$
\State Update trainable parameters using Eq.~\ref{eq:loss}, target $a-z_c$, and true mask $M$
\Statex \textbf{Inference}
\State Encode $z_s\gets\mathcal E(X)$; set $z^{(0)}\gets z_s$
\State Construct descending schedule $\tau_0=1>\cdots>\tau_K=0$
\For{$k=0,\ldots,K-1$}
  \State Evaluate guided velocity $\widehat v_\theta$ with fixed source and mask
  \State $z^{(k+1)}\gets z^{(k)}+(\tau_{k+1}-\tau_k)\widehat v_\theta$
\EndFor
\State \Return $\mathcal D(z^{(K)})$
\end{algorithmic}
\end{algorithm}

\section{Euler Error and Endpoint Prediction}
\label{app:derivation}

Repeated updates change both the states seen by the predictor and the numerical approximation of its flow. For a known training pair, the displacement in Eq.~\ref{eq:velocity} is constant, and an oracle reaches the clean latent from the perturbed source endpoint in one step. The network estimates this displacement from the current state and time, allowing its prediction to change along the trajectory.

Consider the learned guided field $f(z,\tau)=\widehat v_\theta(z,\tau,\vc;z_s,M_{\rm model})$ with fixed conditions, and let $z_f(\tau)$ be its exact ODE solution initialized at $z_s$. If $f$ is Lipschitz in $z$ and sufficiently smooth along this solution, Euler integration on a fixed interval has global error of order $h$, where $h=\max_k|\Delta\tau_k|$. The endpoint reconstruction error can be decomposed as
\begin{equation}
 \|z^{(K)}-z_c\|
 \leq \underbrace{\|z^{(K)}-z_f(0)\|}_{\text{integration error}}
 +\underbrace{\|z_f(0)-z_c\|}_{\text{learned-flow endpoint error}}.
 \label{eq:error}
\end{equation}
Reducing the maximum step size $h$ tightens the integration-error bound under these assumptions. Endpoint accuracy also depends on the learned field and uncertainty about hidden content, so integration accuracy alone does not establish improved detail. The following derivation gives the error recursion and explains the state-dependent predictor. The shared-checkpoint study in Section~\ref{sec:step_study} tests reconstruction quality directly as $K$ varies.

To derive the standard Euler approximation bound for the learned field, set $u=1-\tau$ and $F(z,u)=-f(z,1-u)$, so integration proceeds forward from $u=0$ to $u=1$. Assume $F$ is Lipschitz in $z$ with constant $L$ on a region containing the exact and numerical trajectories, and that the exact solution satisfies $\|d^2z_F/du^2\|\le B$ there. Taylor expansion over an interval $h_k$ then gives
\begin{equation}
 z_F(u_{k+1})=z_F(u_k)+h_k F(z_F(u_k),u_k)+r_k,
 \qquad \|r_k\|\le \tfrac12 B h_k^2.
\end{equation}
Subtracting this update from Euler and using the Lipschitz condition yields
\begin{equation}
 e_{k+1}\le(1+Lh_k)e_k+\tfrac12 B h_k^2,
 \qquad e_k=\|z^{(k)}-z_F(u_k)\|.
\end{equation}
With the same initial condition, $e_0=0$, and setting $h=\max_k h_k$ gives the iterated bound, for $L>0$,
\begin{equation}
 e_K\le \frac{Bh}{2L}(e^L-1),
\end{equation}
and the $L=0$ limit is $Bh/2$. Here, $L$ and $B$ are assumed constants rather than measured properties of the trained network, and the bound concerns approximation of its ODE rather than the distance between the ODE endpoint and the clean target.

\paragraph{Why a straight paired path can yield a varying predictor.}
The population minimizer of an unweighted squared velocity loss is the conditional mean
\begin{equation}
 v^*(z,\tau,\xi)=\E[a-z_c\mid z_\tau=z,\tau,\xi],
 \qquad \xi=(z_s,M_{\rm model},\vc).
\end{equation}
For the coordinate-weighted loss in Eq.~\ref{eq:loss}, the corresponding minimizer is the weighted conditional mean, $\E[w_i(a_i-z_{c,i})\mid z,\tau,\xi]/\E[w_i\mid z,\tau,\xi]$, when the denominator is positive and the coordinate's valid-sample weighting is absorbed into $w_i$. Although a particular paired displacement is constant, the conditional distribution over displacements can depend on the observed intermediate state, which explains why evaluating the field again can change the predicted correction. This dependence does not ensure faithful recovery of a multimodal hidden background, an outcome tested directly by the matched-checkpoint experiment.

\section{Metric Definitions and Reporting Protocol}
\label{app:metrics}
Tables~\ref{tab:main} and~\ref{tab:categories} use the complete 1,631-video release and pixel-accurate \texttt{mask\_packed} GT masks, with scores transcribed from the released machine-readable records. Except for global VFID and the aggregate TWE gap, frame or frame-pair measurements are averaged within videos and then equally across videos. For the finite PSNR mean, infinite-valued video scores are omitted and counted separately.

\paragraph{Pixel and structural fidelity.}
Mask PSNR is computed from errors over GT mask pixels in each frame, while Mask SSIM averages the spatial SSIM map over those pixels rather than over the enclosing rectangle. Input masks and scoring masks are separate, so a pipeline may receive bounding boxes while its output is scored on the common pixel mask.

\paragraph{Perceptual fidelity.}
For LPIPS and DISTS, form $\widehat Y_t^{M}=M_t\odot\widehat Y_t+(1-M_t)\odot Y_t$ to replace the exterior prediction with clean-target pixels, then use a temporally stable crop enclosing the union of mask locations over the video for both prediction and target. Active-mask frames are resized to a short side of 256 pixels with the long side capped at 512. For VFID, the same GT-composited crop video supplies RGB-I3D features of dimension 1,024 through the ProPainter evaluator, and the global Fr\'echet distance is computed between feature distributions rather than averaged from per-video or per-type distances. These constructions isolate errors around the overlay and do not measure exterior preservation.

\paragraph{Released temporal diagnostics.}
Optical flow uses Farneback with maximum image side 320, and the temporal region is the union of the current GT mask and the previous mask transported to the current frame by GT backward flow. Flow statistics instead use the previous mask union the current mask sampled into previous-frame coordinates by GT forward flow. TWE uses the sampling operation defined in Appendix~\ref{app:temporal_diagnostics} and restricts error aggregation to the mask region, with RGB errors on the $[0,255]$ scale. TC, Flow Mean, and Flow Var retain the release's named diagnostics and numerical scale, as reported in Table~\ref{tab:main} using this released computation.

The released TWE gap is computed \emph{after} aggregation:
\begin{equation}
 \Delta_{\rm TWE}=\left|\frac{1}{N}\sum_{i=1}^N\mathrm{TWE}(\widehat Y_i)
 -\frac{1}{N}\sum_{i=1}^N\mathrm{TWE}(Y_i)\right|,\qquad N=1{,}631.
\end{equation}
With a common target reference of 9.3228, this dataset-level statistic accompanies the target-referenced reconstruction metrics, and Appendix~\ref{app:temporal_diagnostics} specifies its interpretation.

\section{Overlay Categories and Evaluation Coverage}
\label{app:categories}
\begin{table}[ht]
\caption{\textbf{DVTE-Bench coverage and complete evaluation.} All released clips are evaluated by five methods under the protocol defined in Section~\ref{sec:benchmark_protocol}.}
\label{tab:benchmark_coverage}
\centering\small
\setlength{\tabcolsep}{4pt}
\begin{tabular}{llrr}
\toprule
Challenge type & Overlay characteristic & Release & Evaluated\\
\midrule
Standard subtitle & Lower-third captions and style variation & 232 & 232\\
Artistic title & Display typography, outlines, and glow & 218 & 218\\
Moving text & Frame-varying trajectories & 232 & 232\\
Vertical text & Vertical writing at varied positions & 215 & 215\\
Non-bottom layout & Placement outside the bottom region & 223 & 223\\
Multi-region & Distributed simultaneous text regions & 225 & 225\\
Persistent watermark & Persistent semi-transparent text marks & 218 & 218\\
ASR subtitle & Speech-aligned captions & 68 & 68\\
\midrule
Full release & & 1,631 & 1,631\\
\bottomrule
\end{tabular}
\end{table}

\begin{figure}[ht]
\centering
\includegraphics[width=\linewidth]{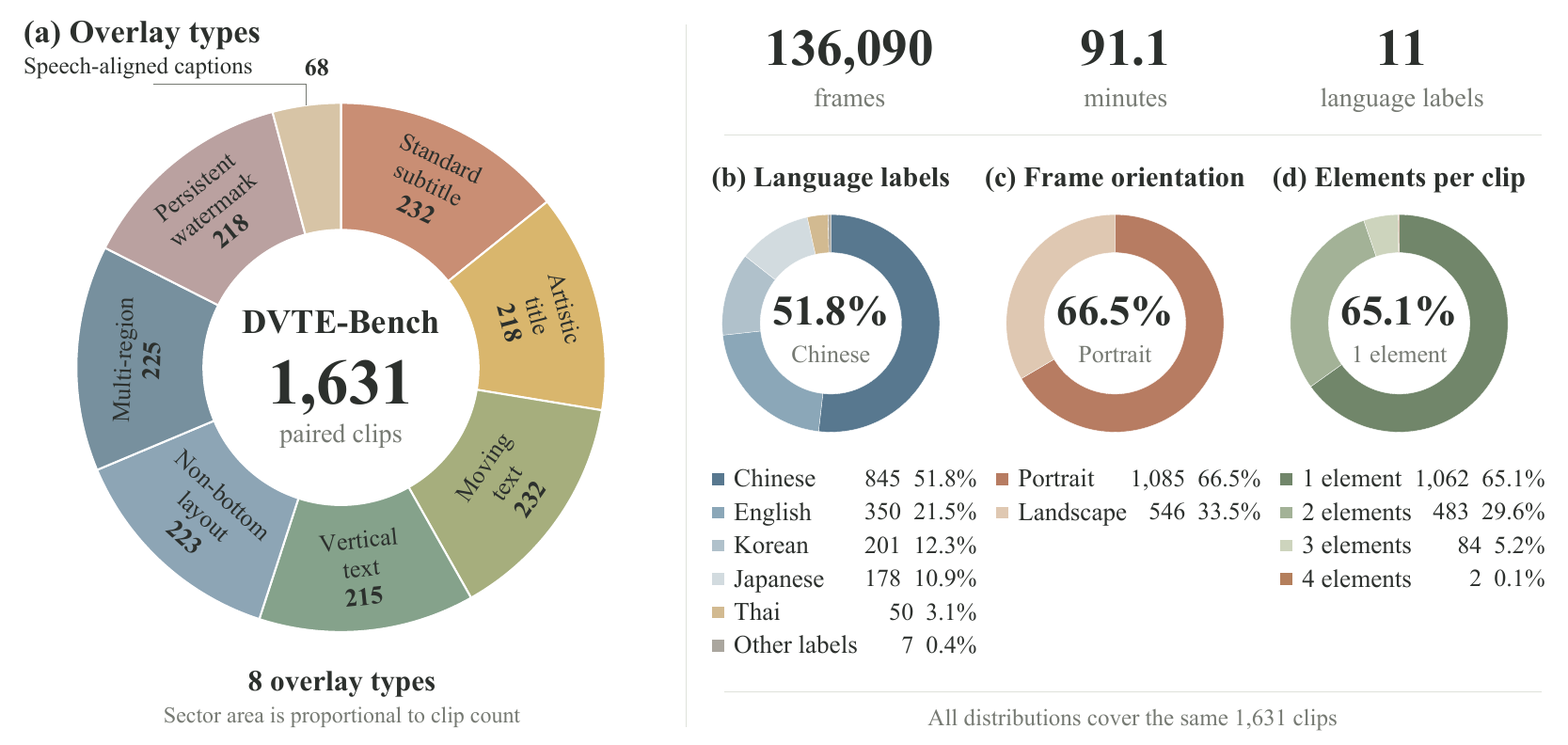}
\caption{\textbf{DVTE-Bench release distributions.} All panels count the 1,631 released clips: (a) assigned overlay type, (b) language metadata, (c) frame orientation, and (d) the per-clip number of rendered overlay elements. ``Other labels'' combines Italian (2) and mixed, Spanish, Portuguese, Polish, and Welsh (1 each). Overlay-element count does not measure spatial regions or pixel coverage.}
\label{fig:benchmark_distribution}
\end{figure}

Overlay function and benchmark challenge type describe different properties: subtitles, text watermarks, titles, and name or role captions describe function, while position, motion, opacity, and affected area describe appearance and layout. A vertical name caption, for example, combines a function with a spatial layout, although the current benchmark assigns each clip one challenge type. Its clean-target OCR policy excludes confirmed text, so preservation of natural scene text requires a separate evaluation setting.

\paragraph{Dataset composition.}
The release contains 1,631 clips across eight overlay types, evaluated under the protocol defined in Section~\ref{sec:benchmark_protocol}. Figure~\ref{fig:benchmark_distribution} summarizes the manifest counts, including 1,085 portrait and 546 landscape clips. Language metadata records Chinese (845), English (350), Korean (201), Japanese (178), Thai (50), Italian (2), and one clip each labeled mixed, Spanish, Portuguese, Polish, and Welsh. These counts describe coverage, but the rare labels do not support reliable per-language comparisons. Every target frame is scanned under the OCR rejection policy described in Section~\ref{sec:benchmark}.

DVTE-Bench organizes overlay conditions by layout, motion, typography, and caption timing, with Table~\ref{tab:categories} reporting \method{} across all eight types and 1,631 clips. Aggregates weight videos equally rather than assigning equal weight to each type, and \method{} leads Mask PSNR, SSIM, LPIPS, and DISTS among the five evaluated methods in each type.

\begin{table}[ht]
\caption{\textbf{DeOverlay across all eight DVTE-Bench types.} All 1,631 clips are included, with the 40K checkpoint using GT bounding-box masks, six steps, and the released seamless paste-back setting. Scoring uses pixel masks, and $N_{\infty}$ counts clips omitted only from finite-mean PSNR.}
\label{tab:categories}
\centering\small
\setlength{\tabcolsep}{3.5pt}
\begin{tabular*}{\linewidth}{@{\extracolsep{\fill}}lrrrrrr@{}}
\toprule
Text type & Clips & PSNR$^{\dagger}\uparrow$ & SSIM$\uparrow$ & LPIPS$\downarrow$ & DISTS$\downarrow$ & $N_{\infty}$\\
\midrule
Standard subtitle & 232 & 30.39 & 0.8718 & 0.0285 & 0.0395 & 1\\
Artistic title & 218 & 26.24 & 0.8409 & 0.0674 & 0.0668 & 0\\
Moving text & 232 & 30.73 & 0.8800 & 0.0209 & 0.0325 & 4\\
Vertical text & 215 & 30.46 & 0.8736 & 0.0975 & 0.1015 & 3\\
Non-bottom layout & 223 & 30.92 & 0.8843 & 0.0414 & 0.0628 & 2\\
Multi-region & 225 & 29.88 & 0.8834 & 0.0103 & 0.0167 & 2\\
Persistent watermark & 218 & 36.02 & 0.9306 & 0.0314 & 0.0380 & 3\\
ASR subtitle & 68 & 34.20 & 0.9184 & 0.0137 & 0.0527 & 0\\
\bottomrule
\end{tabular*}
\end{table}

\paragraph{Evaluated pipelines.}
The benchmark metadata identifies the 40K checkpoint, GT bounding-box masks, six steps, and a 12-pixel seamless paste-back setting for \method{}. ProPainter, MiniMax-Remover, and DiffuEraser share GT bounding-box masks and copy source pixels outside an 8-pixel dilated mask, while CLEAR uses official mask-free inference. These settings define the evaluated system pipelines, with the dataset GT mask serving as the scoring region for every method. Appendix~\ref{app:metrics} specifies the pixel-mask scoring and perceptual crop protocol, while the qualitative exports in Appendix~\ref{app:qualitative_exports} retain their own run provenance.

Mask provenance is recorded independently of category, distinguishing reconstruction with a known edit region under an oracle mask from evaluation of the associated detection pipeline under an estimated mask. A manually supplied mask is a separate input setting and will not be labeled automatic detection. For watermarks and captions with graphic elements, coverage must include the entire designated overlay rather than only recognized characters.

\section{Qualitative Exports and Case Analysis}
\label{app:qualitative_exports}
The application examples in Figure~\ref{fig:qualitative} display matched source and \method{} output frames at the same frame index, with the source on the left and the output on the right in each pair.

\begin{figure}[htbp]
\centering
\includegraphics[page=1,width=\linewidth]{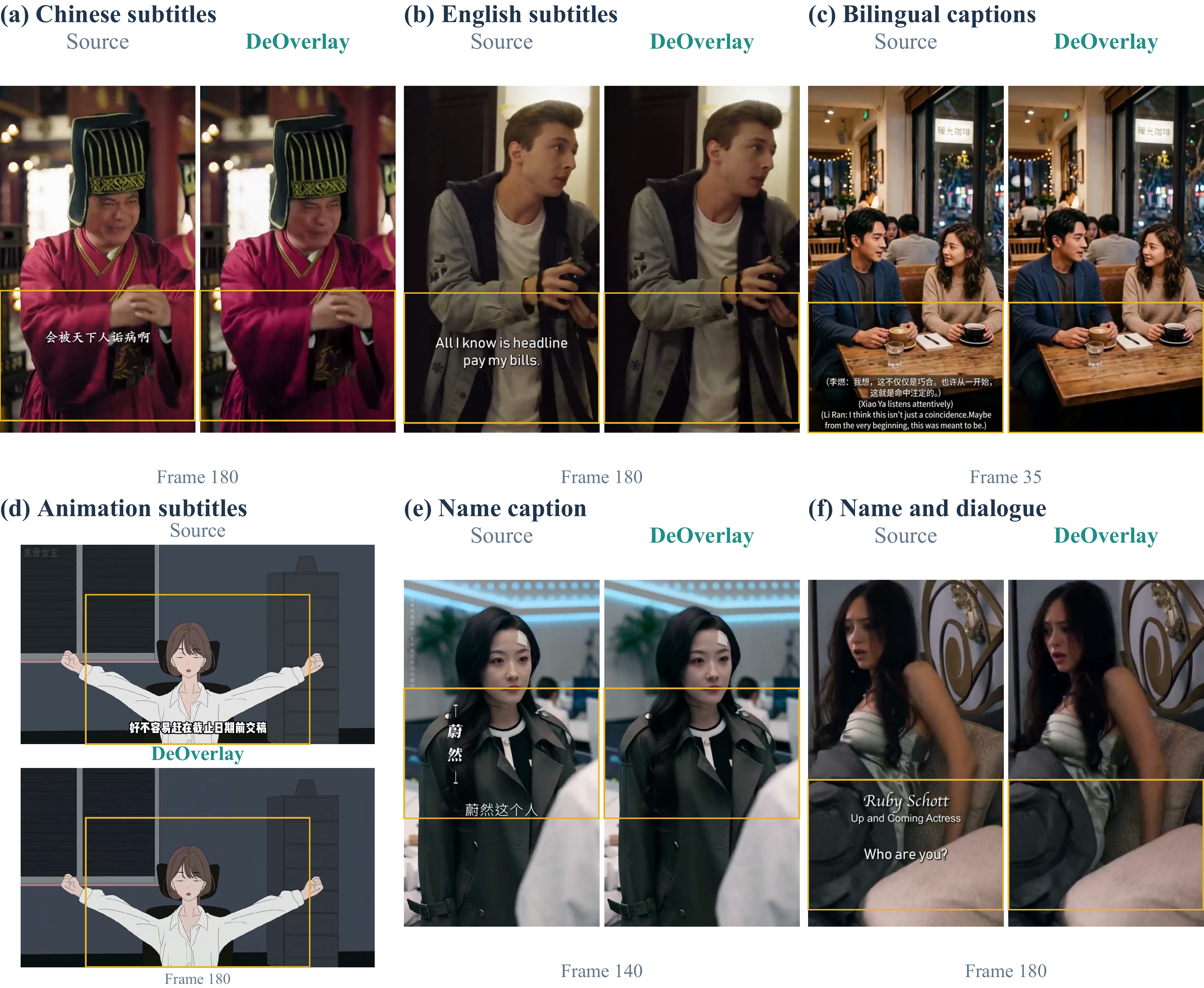}
\caption{\textbf{Application context for Figure~\ref{fig:qualitative}(a--f).} Adjacent source/output frames preserve native orientation, with yellow boxes locating the matching main-paper crops and zero-based indices identifying the frames.}
\label{fig:qualitative_full_a}
\end{figure}

\begin{figure}[htbp]
\centering
\includegraphics[page=2,width=\linewidth]{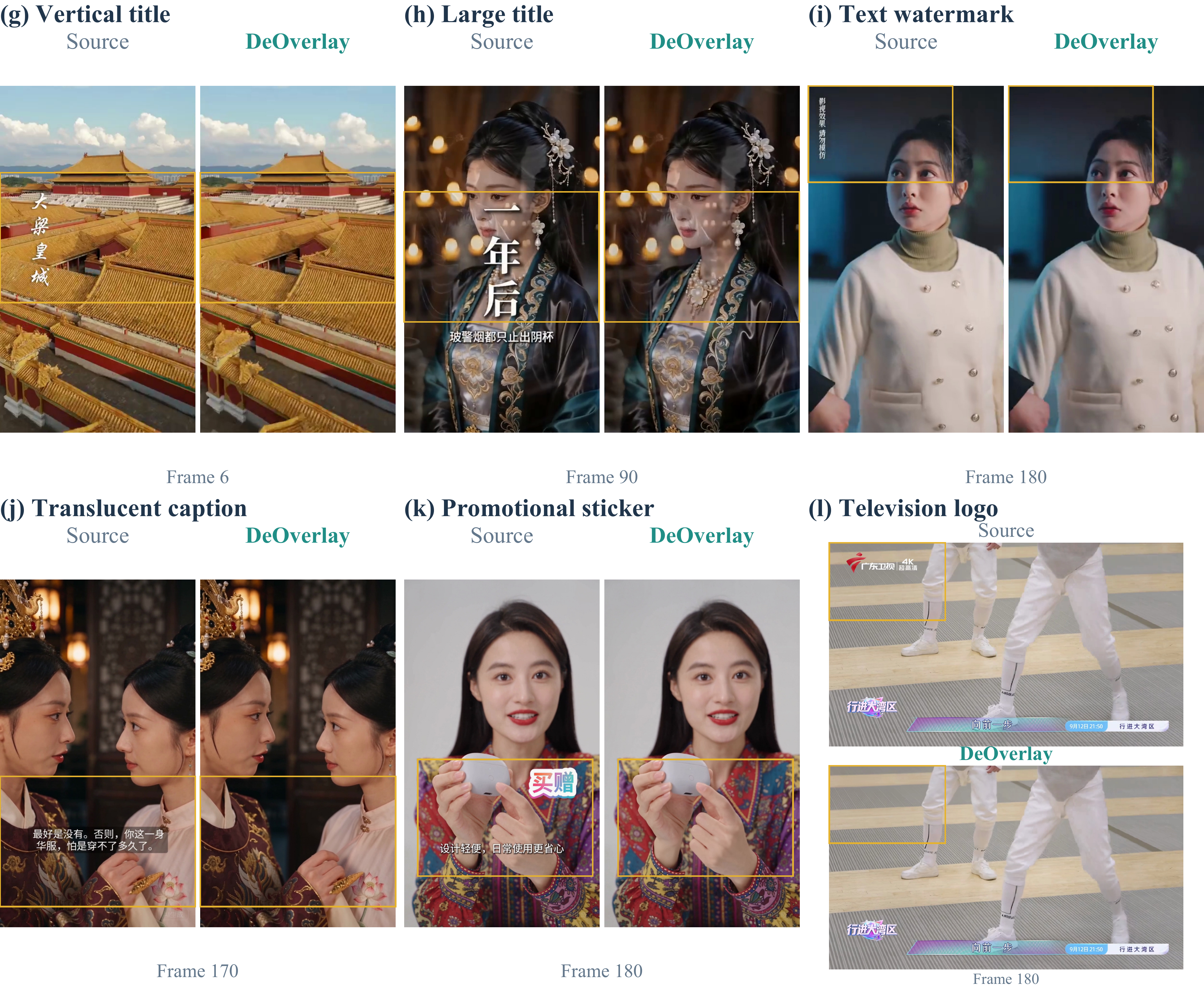}
\caption{\textbf{Application context for Figure~\ref{fig:qualitative}(g--l).} Full source/output pairs include titles, watermarks, captions, a sticker, and the selected television-logo edit, which retains unrelated program graphics in the logo-only case. Yellow boxes locate the main-paper crops.}
\label{fig:qualitative_full_b}
\end{figure}

\begin{figure}[htbp]
\centering
\includegraphics[width=\linewidth]{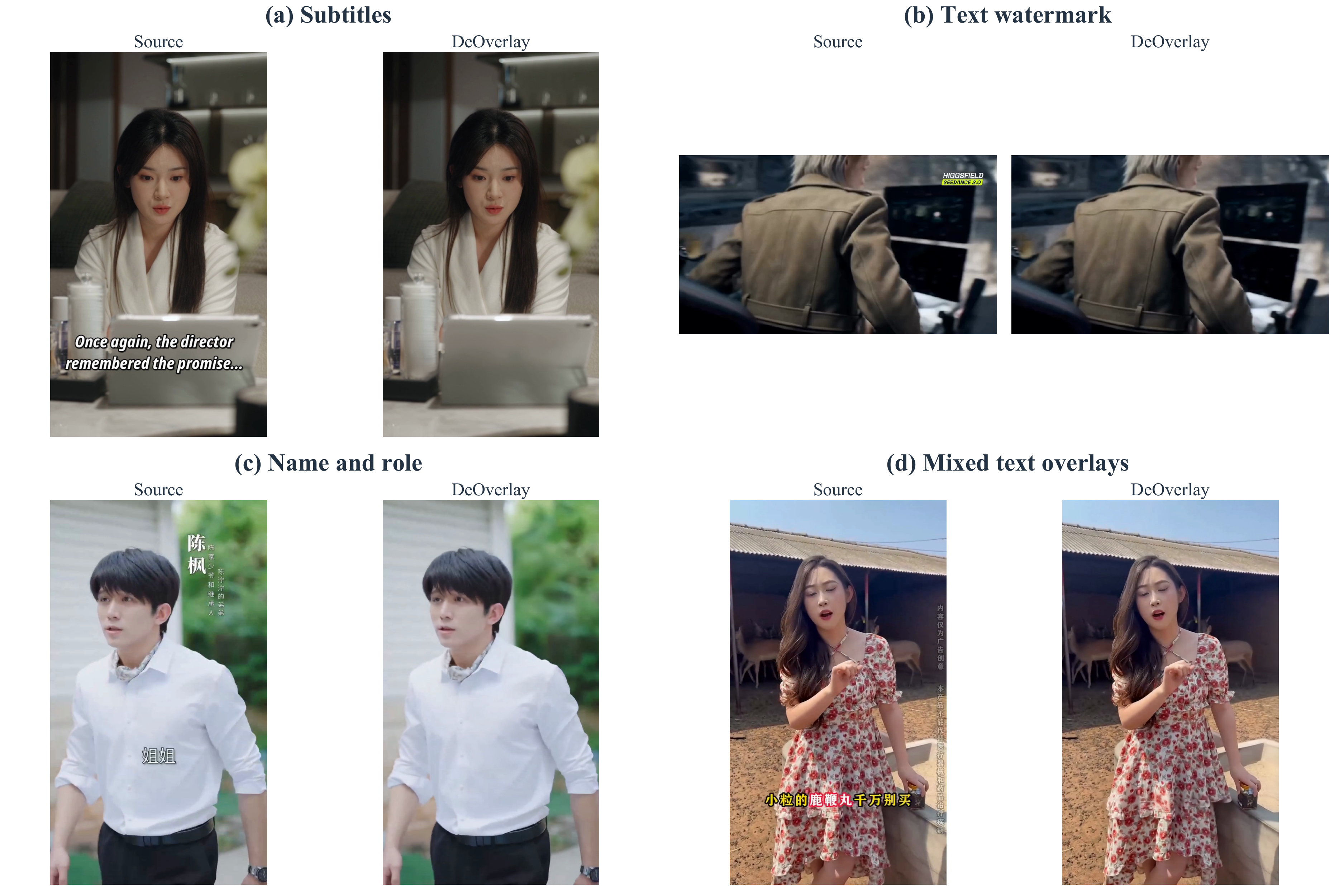}
\caption{\textbf{Additional source/output examples.} Paired full frames show subtitles, a text watermark, a name caption, and mixed text overlays from the earlier export collection, with native orientations preserved for these application clips without clean targets.}
\label{fig:qual_context}
\end{figure}

\paragraph{Selection and presentation.}
Figure~\ref{fig:qualitative} uses twelve matched source/output crops from the supplied current-model exports, covering Chinese, English, bilingual, and animation subtitles, as well as name captions and dialogue, vertical and large titles, a text watermark, a translucent caption, a promotional sticker, and a television logo. Figures~\ref{fig:qualitative_full_a} and~\ref{fig:qualitative_full_b} retain the corresponding full frames and mark the crop coordinates, which are also recorded with frame indices in the figure-source manifest and editable figure notes. Figure~\ref{fig:qual_context} retains four additional source/output examples from the earlier collection. The television-logo case uses a logo-only mask that leaves the remaining program graphics outside the requested edit, while the large-title source already contains a transition double exposure visible on both sides. These exports are not a controlled step-count comparison, and their presentation uses frame extraction, equal-coordinate cropping, and display scaling with no color or contrast enhancement. Across the broader export collection, occasional subtitle remnants, local texture artifacts, and residual geometric watermark elements remain.

\begin{figure}[htbp]
\centering
\includegraphics[page=2,width=\linewidth]{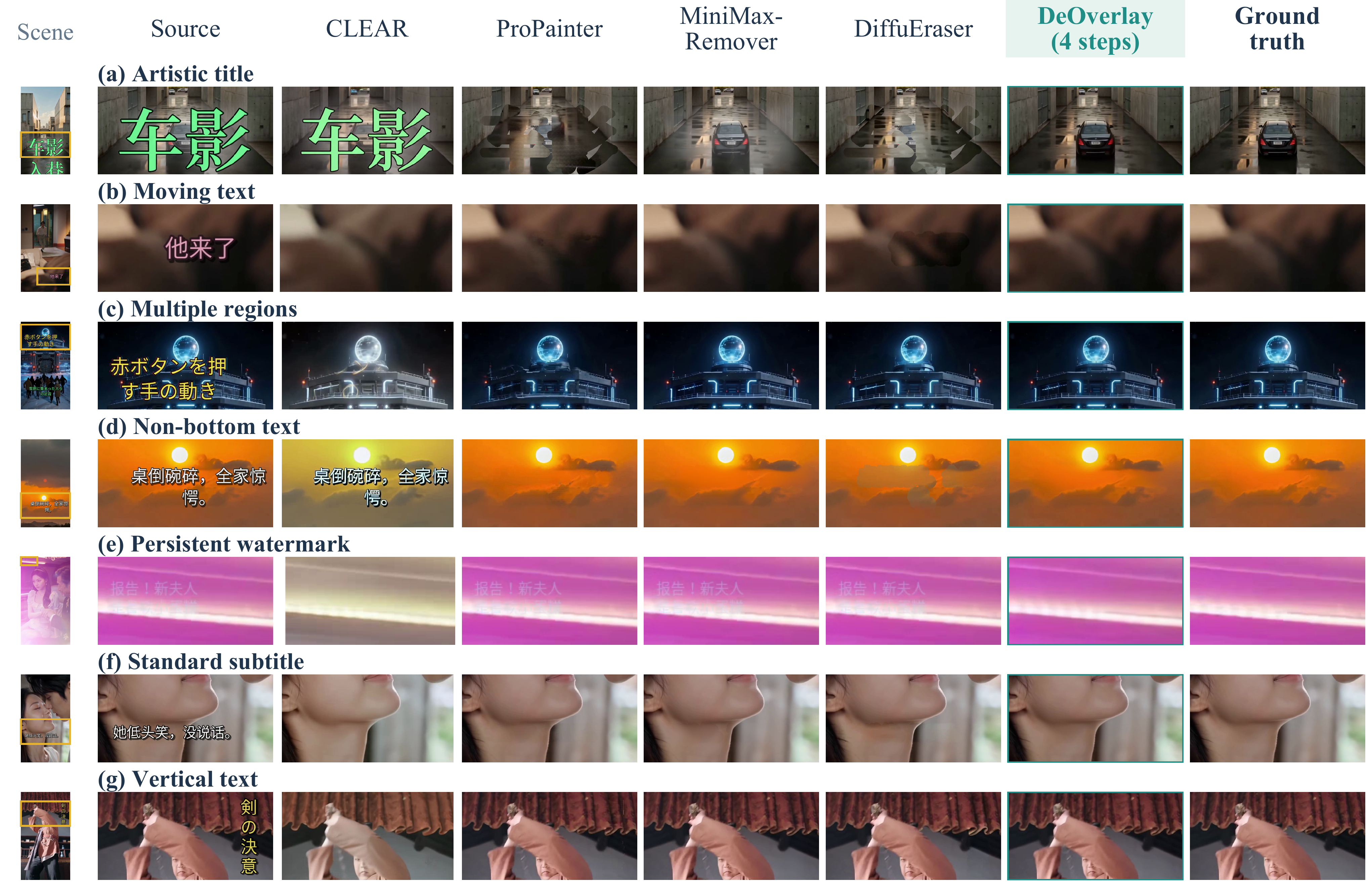}
\caption{\textbf{Expanded comparison across seven overlay categories.} Every row shows an identical frame and common crop for the source, four baselines, \method{} ($K=4$), and clean target, with yellow boxes locating the crops in small source frames.}
\label{fig:benchmark_comparison_all}
\end{figure}

\begin{figure}[htbp]
\centering
\includegraphics[page=3,width=\linewidth]{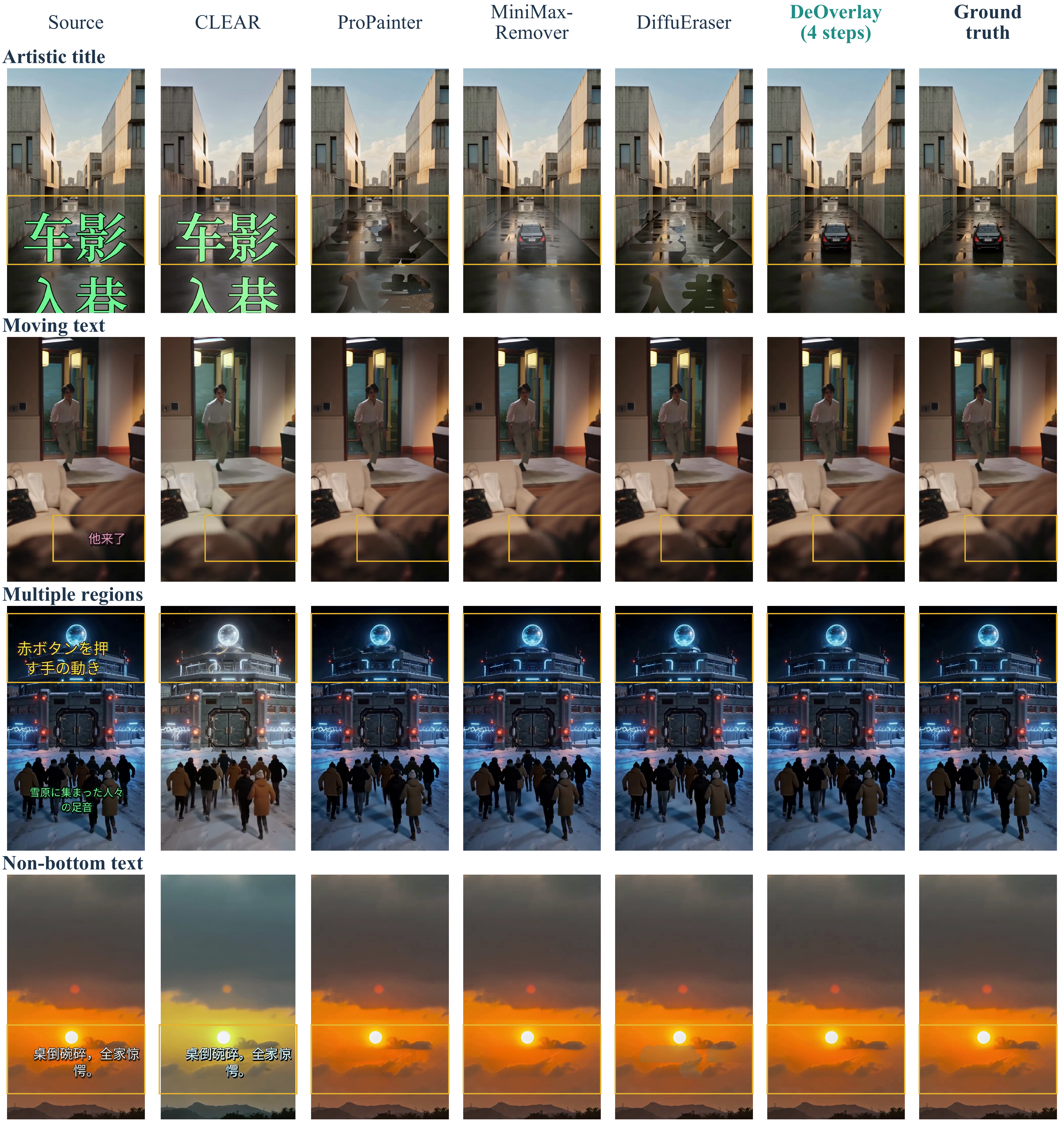}
\caption{\textbf{Benchmark context: titles, moving text, and spatial layouts.} Each row uses the same clip and frame for all methods, with yellow boxes marking the source-coordinate crops in Figures~\ref{fig:benchmark_comparison} and~\ref{fig:benchmark_comparison_all}.}
\label{fig:benchmark_comparison_context}
\end{figure}

\begin{figure}[htbp]
\centering
\includegraphics[page=4,width=\linewidth]{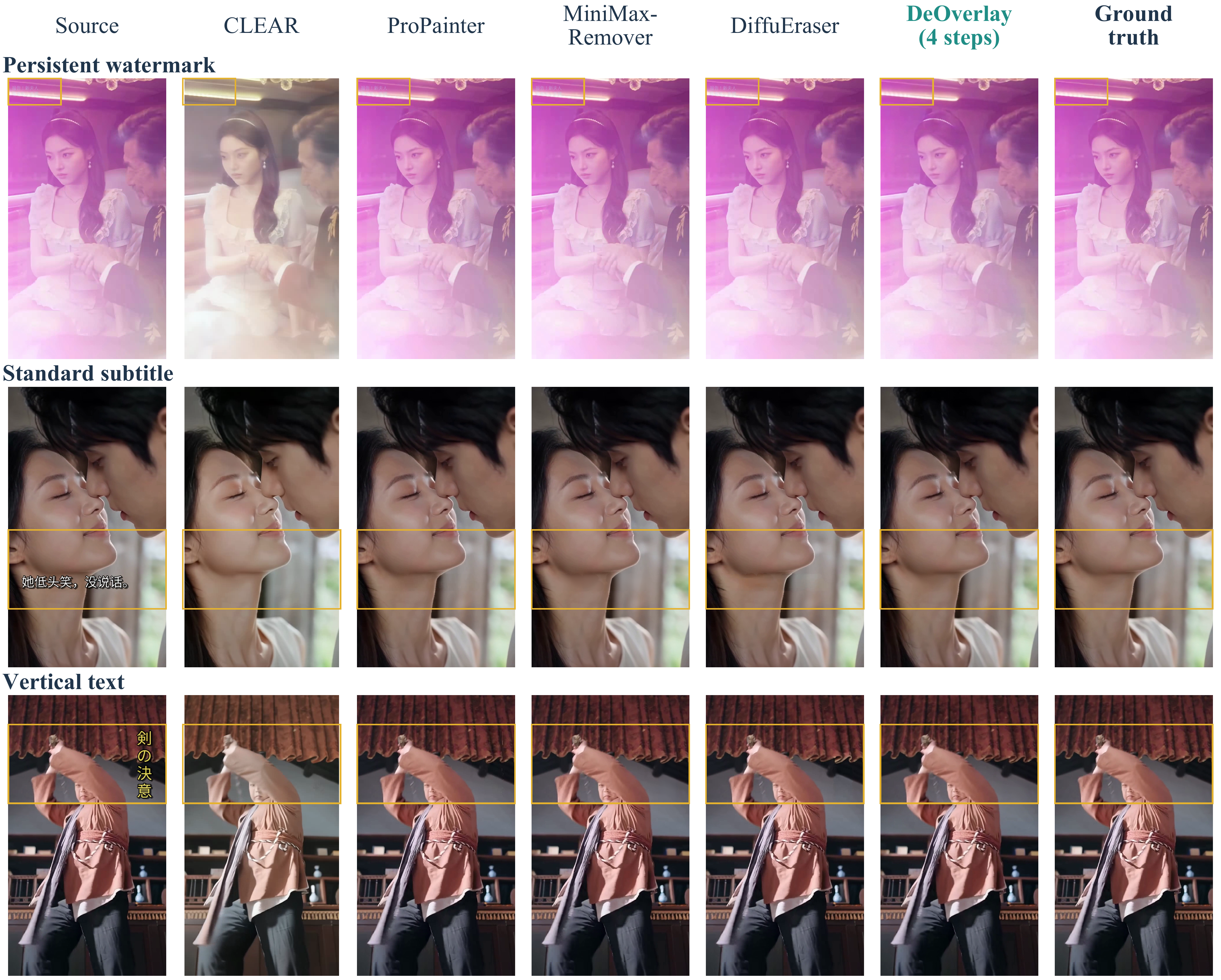}
\caption{\textbf{Benchmark context: watermarks, subtitles, and vertical text.} Full frames retain the surrounding scene and all overlaid regions, while yellow boxes locate the identical crops in Figure~\ref{fig:benchmark_comparison_all}.}
\label{fig:benchmark_comparison_context_b}
\end{figure}

\paragraph{Paired benchmark context.}
Figure~\ref{fig:benchmark_comparison_all} extends the five main-paper examples to seven overlay categories by adding a persistent watermark and a standard subtitle, with all seven cases containing 81 frames. Figures~\ref{fig:benchmark_comparison_context} and~\ref{fig:benchmark_comparison_context_b} show the corresponding full frames for every method, using zero-based frame 40 and the four-step DeOverlay outputs from the same-A100 timing package. CLEAR's native output is fitted without changing its aspect ratio and centered on the source-sized canvas before applying the common crop. The artistic-title crop focuses on the upper title and car, while the multi-region crop focuses on the upper overlay and leaves the second region visible in the full frame. The small, translucent watermark is enlarged without contrast adjustment. These selected cases illustrate restoration behavior rather than estimating an aggregate ranking or comparing inference depths.

\begin{figure}[htbp]
\centering
\includegraphics[width=\linewidth]{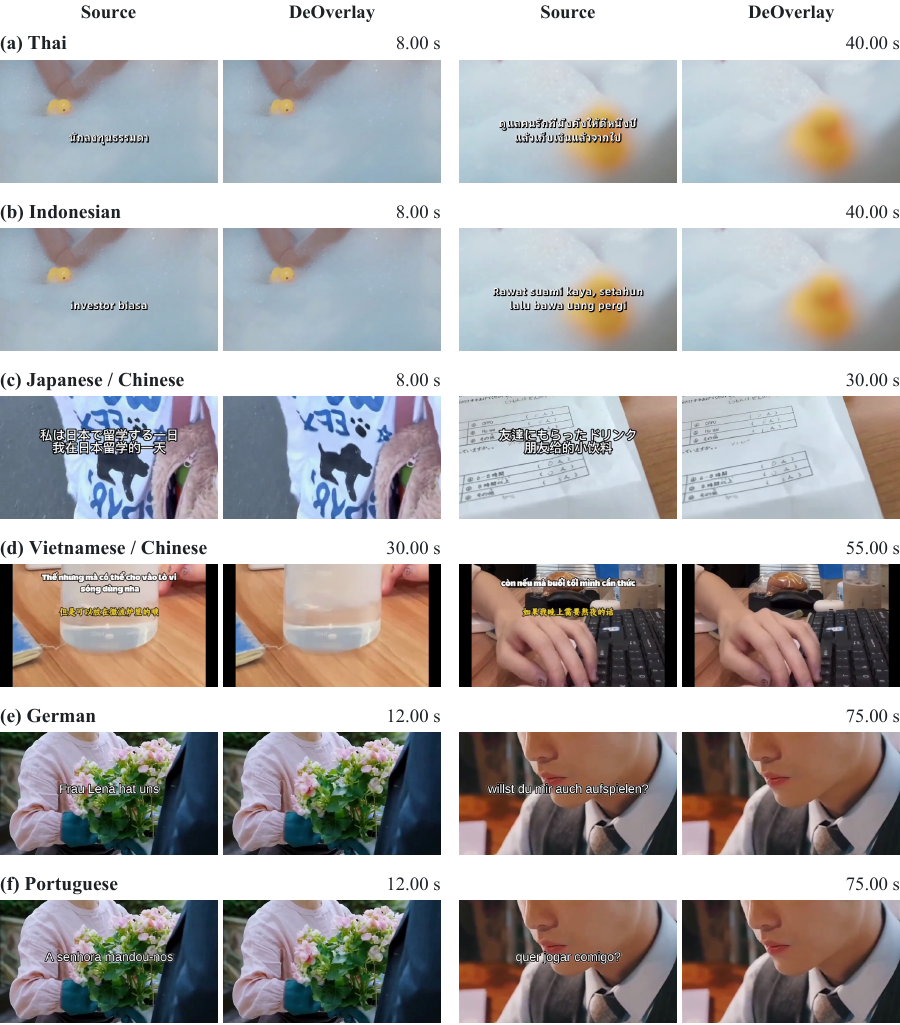}
\caption{\textbf{Multilingual generalization on application videos.} Six language settings include Thai, Indonesian, Japanese--Chinese, Vietnamese--Chinese, German, and Portuguese overlays. Each row contains two matched source/\method{} pairs, with timestamps identifying the displayed frames. For these application clips without clean targets, input and output use identical crop coordinates and display scales.}
\label{fig:qual_multilingual}
\end{figure}

\paragraph{Multilingual generalization.}
Figure~\ref{fig:qual_multilingual} shows Thai, Indonesian, German, Portuguese, Japanese--Chinese, and Vietnamese--Chinese text overlays. The Thai and Indonesian clips share the same scene content, as do the German and Portuguese clips, making differences in glyph shapes and line layouts visible on matching backgrounds, while the bilingual examples combine two writing systems within the same frame. Each row includes two time points with matched source/output coordinates, showing reconstruction beneath changing subtitle content and across different scenes, with background patterns, clothing, printed scene text, and objects remaining visible in the outputs.

\FloatBarrier
\section{Ablation Details}
\label{app:mechanism_studies}
The component and step-count ablations use 200 randomly sampled videos from DVTE-Bench, while the main comparison in Table~\ref{tab:main} evaluates all 1,631 released videos. The step sweep in Table~\ref{tab:steps} holds the multistep checkpoint fixed and changes the inference update count, whereas Table~\ref{tab:core_ablation} uses four steps for the component variants and reports a separately trained, matched one-step direct-residual reference. The quality tables retain the reported Mask MAE, Outside MAE, MaskCrop SSIM, and TWE absolute-gap scales.

\subsection{Training Formulation and Initialization}
\label{sec:detail_diagnosis}
Under the same experimental environment, we independently train a one-step direct-residual baseline to predict the source-to-target residual. In Table~\ref{tab:core_ablation}, four-step flow improves Mask PSNR from 31.300 to 32.461 dB, reduces MAE from 4.520 to 4.299, raises Crop SSIM from 0.91120 to 0.92765, and reduces TWE absolute gap from 0.195 to 0.180. These results compare the two trained models at their respective inference settings, while Table~\ref{tab:steps} studies update count within the continuous-flow checkpoint.

The Gaussian-start model in Table~\ref{tab:core_ablation} is trained for a noise-initialized, two-video-stream input, so its 2.861 dB gap to the source-initialized model reflects both the starting state and the way source evidence enters the network. The independent token microbenchmark below evaluates the computational effect of the second stream.

\subsection{Source Conditioning}
\label{sec:guidance_ablation}
The full four-step model has Outside MAE of 2.541, rising to 2.552 with both source branches removed, 2.543 without VSA, and 2.550 without GSC. These small increases accompany the covered-region changes in Table~\ref{tab:core_ablation} and describe exterior behavior within the component study. Keeping both branches but replacing visibility-normalized aggregation with an ordinary temporal mean gives Outside MAE of 2.542 and a TWE absolute gap of 0.183, compared with 0.180 for the full model.

\subsection{Transformer Token Cost}
\begin{table}[htbp]
\caption{\textbf{One versus two video-token streams.} The microbenchmark uses real 32K weights in bfloat16 and a latent tensor of shape $1\times16\times23\times36\times64$, corresponding to an 81-frame input with eight mirror-padding frames at the 768s bucket. Both inputs include 98 text tokens, and timing covers four transformer forwards while excluding VAE encoding, decoding, and video processing. Memory is incremental peak allocation above the loaded runtime.}
\label{tab:token_cost}
\centering\small
\setlength{\tabcolsep}{4pt}
\begin{tabular*}{\linewidth}{@{\extracolsep{\fill}}lrrrr@{}}
\toprule
Input organization & Video tokens & Total tokens & Time (s)$\downarrow$ & Memory (GiB)$\downarrow$\\
\midrule
Source-initialized stream & 13,248 & 13,346 & 5.266 & 1.744\\
Noise and source streams & 26,496 & 26,594 & 13.088 & 3.479\\
\bottomrule
\end{tabular*}
\end{table}

We warm up each input organization and interleave four timed trials, reversing their order between rounds and bounding each measurement with CUDA synchronization. The table reports medians, with 10th--90th percentile time ranges of 5.260--5.272 s and 13.077--13.099 s, respectively, while incremental memory subtracts the pre-forward allocation from the peak allocation. The machine can carry concurrent GPU load, so these timings characterize the recorded workload rather than isolated end-to-end deployment speed.

\section{Timing and Optimization Protocol}
\label{app:reporting_details}
\paragraph{Shared input workload.}
The end-to-end comparison in Table~\ref{tab:runtime} uses eight deterministically selected clips, one per overlay type. Seven clips contain 81 frames at $540\times960$ resolution, and the remaining clip contains 83 frames at $876\times492$, yielding 650 input frames. Every method processes these same inputs in serial order on one otherwise idle A100 while retaining its internal resizing and processing operations. MiniMax-Remover uses six steps with maximum side 960, whereas DiffuEraser uses maximum side 960, model offloading, and maximum attention slicing.

\paragraph{Timing boundaries and aggregation.}
The per-video measurements exclude one-time model loading, while end-to-end time includes each pipeline's decoding, inference, postprocessing, and video encoding. Seconds per frame is the sum of measured seconds divided by 650, and FPS is its reciprocal. Because the inference-only intervals follow each runner's CUDA synchronization boundaries, end-to-end time is the primary cross-method comparison. For \method{}, the logged inference interval includes the diffusion stage and its associated codec operations, rather than transformer forwards alone.

\paragraph{Memory measurement.}
A common external sampler queries GPU memory every 200 ms across each method's process, and Table~\ref{tab:runtime} reports the largest sampled value in GiB. The one- and four-step \method{} runs share one process and therefore share its recorded peak of 12.236 GiB, which includes the loaded runtime and differs from the incremental forward-allocation measurements in Table~\ref{tab:token_cost}.

\paragraph{Optimization.}
The configured optimizer is AdamW with learning rate $5\times10^{-5}$, betas $(0.9,0.95)$, zero weight decay, 300 warm-up steps, and gradient clipping at 1.0, using bfloat16 and a maximum of 60,000 updates.

\clearpage
\section{Complete Per-Category Comparison}
\label{app:category_comparison}
Tables~\ref{tab:category_comparison_a} and~\ref{tab:category_comparison_b} give the complete five-method comparison for each overlay type under the protocol in Appendix~\ref{app:categories}.
\begin{table}[htbp]
\caption{\textbf{Reconstruction across overlay categories (1/2).} All five methods are evaluated on every clip in each category under the same pixel-mask scoring protocol, with DeOverlay using the 40K checkpoint and six steps. PSNR is the finite mean, $N_{\infty}$ counts infinite-valued video scores, and bold marks the best quality score within each category.}
\label{tab:category_comparison_a}
\centering\small
\setlength{\tabcolsep}{5pt}
\begin{tabular*}{\linewidth}{@{\extracolsep{\fill}}lrrrrr@{}}
\toprule
Method & PSNR$\uparrow$ & SSIM$\uparrow$ & LPIPS$\downarrow$ & DISTS$\downarrow$ & $N_{\infty}$\\
\midrule
\multicolumn{6}{l}{\textit{Standard subtitle (232 clips)}}\\
CLEAR & 17.98 & 0.5726 & 0.2192 & 0.1567 & 0\\
ProPainter & 25.17 & 0.7668 & 0.0890 & 0.0778 & 1\\
MiniMax-Remover & 24.92 & 0.7752 & 0.0838 & 0.0755 & 0\\
DiffuEraser & 24.17 & 0.7501 & 0.0899 & 0.0783 & 1\\
\method{} ($K=6$) & \textbf{30.39} & \textbf{0.8718} & \textbf{0.0285} & \textbf{0.0395} & 1\\
\midrule
\multicolumn{6}{l}{\textit{Artistic title (218 clips)}}\\
CLEAR & 8.67 & 0.2878 & 0.5548 & 0.3276 & 0\\
ProPainter & 19.20 & 0.6639 & 0.2766 & 0.1707 & 0\\
MiniMax-Remover & 22.16 & 0.7395 & 0.1677 & 0.1149 & 1\\
DiffuEraser & 19.73 & 0.6903 & 0.2078 & 0.1373 & 2\\
\method{} ($K=6$) & \textbf{26.24} & \textbf{0.8409} & \textbf{0.0674} & \textbf{0.0668} & 0\\
\midrule
\multicolumn{6}{l}{\textit{Moving text (232 clips)}}\\
CLEAR & 16.28 & 0.5194 & 0.1890 & 0.1451 & 1\\
ProPainter & 27.61 & 0.8158 & 0.0577 & 0.0569 & 4\\
MiniMax-Remover & 27.16 & 0.8194 & 0.0518 & 0.0556 & 0\\
DiffuEraser & 25.97 & 0.7890 & 0.0566 & 0.0592 & 3\\
\method{} ($K=6$) & \textbf{30.73} & \textbf{0.8800} & \textbf{0.0209} & \textbf{0.0325} & 4\\
\midrule
\multicolumn{6}{l}{\textit{Vertical text (215 clips)}}\\
CLEAR & 13.10 & 0.3324 & 0.5796 & 0.3836 & 0\\
ProPainter & 26.92 & 0.7975 & 0.1756 & 0.1432 & 3\\
MiniMax-Remover & 27.63 & 0.8275 & 0.1604 & 0.1328 & 1\\
DiffuEraser & 26.55 & 0.8023 & 0.1756 & 0.1430 & 4\\
\method{} ($K=6$) & \textbf{30.46} & \textbf{0.8736} & \textbf{0.0975} & \textbf{0.1015} & 3\\
\bottomrule
\end{tabular*}
\end{table}

\begin{table}[htbp]
\caption{\textbf{Reconstruction across overlay categories (2/2).} Settings and notation follow Table~\ref{tab:category_comparison_a}, with the two tables together covering all 1,631 released clips.}
\label{tab:category_comparison_b}
\centering\small
\setlength{\tabcolsep}{5pt}
\begin{tabular*}{\linewidth}{@{\extracolsep{\fill}}lrrrrr@{}}
\toprule
Method & PSNR$\uparrow$ & SSIM$\uparrow$ & LPIPS$\downarrow$ & DISTS$\downarrow$ & $N_{\infty}$\\
\midrule
\multicolumn{6}{l}{\textit{Non-bottom layout (223 clips)}}\\
CLEAR & 17.76 & 0.5941 & 0.3264 & 0.2282 & 0\\
ProPainter & 24.84 & 0.7627 & 0.1462 & 0.1269 & 2\\
MiniMax-Remover & 25.31 & 0.7909 & 0.1195 & 0.1137 & 1\\
DiffuEraser & 24.22 & 0.7582 & 0.1356 & 0.1229 & 2\\
\method{} ($K=6$) & \textbf{30.92} & \textbf{0.8843} & \textbf{0.0414} & \textbf{0.0628} & 2\\
\midrule
\multicolumn{6}{l}{\textit{Multi-region (225 clips)}}\\
CLEAR & 15.81 & 0.5273 & 0.1506 & 0.1017 & 0\\
ProPainter & 23.48 & 0.7634 & 0.0453 & 0.0449 & 2\\
MiniMax-Remover & 24.14 & 0.7915 & 0.0361 & 0.0372 & 1\\
DiffuEraser & 23.10 & 0.7613 & 0.0411 & 0.0401 & 2\\
\method{} ($K=6$) & \textbf{29.88} & \textbf{0.8834} & \textbf{0.0103} & \textbf{0.0167} & 2\\
\midrule
\multicolumn{6}{l}{\textit{Persistent watermark (218 clips)}}\\
CLEAR & 20.96 & 0.7017 & 0.2886 & 0.1835 & 0\\
ProPainter & 33.09 & 0.8858 & 0.0566 & 0.0543 & 3\\
MiniMax-Remover & 30.36 & 0.8570 & 0.0929 & 0.0768 & 0\\
DiffuEraser & 29.23 & 0.8328 & 0.1114 & 0.0888 & 2\\
\method{} ($K=6$) & \textbf{36.02} & \textbf{0.9306} & \textbf{0.0314} & \textbf{0.0380} & 3\\
\midrule
\multicolumn{6}{l}{\textit{ASR subtitle (68 clips)}}\\
CLEAR & 20.50 & 0.7638 & 0.1992 & 0.2097 & 0\\
ProPainter & 29.59 & 0.8698 & 0.0443 & 0.0977 & 0\\
MiniMax-Remover & 31.28 & 0.9064 & 0.0275 & 0.0764 & 0\\
DiffuEraser & 28.06 & 0.8469 & 0.0520 & 0.1025 & 0\\
\method{} ($K=6$) & \textbf{34.20} & \textbf{0.9184} & \textbf{0.0137} & \textbf{0.0527} & 0\\
\bottomrule
\end{tabular*}
\end{table}

\FloatBarrier

\section{Temporal Diagnostic Computation}
\label{app:temporal_diagnostics}
Table~\ref{tab:main}(b) reports the release's temporal measurements on all 1,631 videos, with the TWE label following the forward-flow remap convention specified in the release metadata. Using forward displacement $u_{t-1\to t}$, the referenced computation samples the previous frame as
\begin{equation}
 I_{t-1}^{\rightarrow}(x)=I_{t-1}\!\left(x+u_{t-1\to t}(x)\right)
\end{equation}
and averages its absolute RGB difference from $I_t(x)$ on the scoring region. This forward-displacement sampling diagnostic differs from backward warping of the previous frame into the current frame, which samples with $u_{t\to t-1}$. The table retains the released diagnostic names and numerical scales, including the reported labels for TC, Flow Mean, and Flow Var.

The aggregate gap applies the absolute value after video-level averaging, $|\overline{\mathrm{TWE}}_{\rm pred}-\overline{\mathrm{TWE}}_{\rm GT}|$, with clean-target reference 9.3228, so opposite signed deviations can cancel. Table~\ref{tab:main} also reports the target-referenced reconstruction comparison.

\end{document}